\documentclass[11pt]{article}

\usepackage[preprint]{acl}

\usepackage{times}
\usepackage{latexsym}

\usepackage[T1]{fontenc}

\usepackage[utf8]{inputenc}

\usepackage{microtype}

\usepackage{inconsolata}
\usepackage{rotating}
\usepackage[table]{xcolor}
\usepackage[utf8]{inputenc} 
\usepackage[T1]{fontenc}    
\usepackage{booktabs}       
\usepackage{amsfonts}       
\usepackage{nicefrac}       
\usepackage{microtype}      
\usepackage{xcolor}         
\usepackage[textsize=tiny]{todonotes}
\usepackage{amsmath}
\usepackage{amssymb}
\usepackage{mathtools}
\usepackage{amsthm}
\usepackage{graphicx}
\usepackage{subfigure}
\usepackage{multirow}
\usepackage{tablefootnote}
\usepackage{makecell}
\usepackage{algorithmic}
\usepackage[linesnumbered,ruled]{algorithm2e}
\usepackage{wrapfig}
\usepackage{lipsum,caption}
\usepackage{pifont}
\usepackage{lipsum}
\usepackage{adjustbox}
\usepackage{tabularx}
\usepackage{enumerate}
\usepackage{etoolbox}
\usepackage{diagbox}
\usepackage{threeparttable} 
\usepackage{siunitx}
\usepackage[most]{tcolorbox}
\usepackage{multicol}

\newcommand{\method}{\textsc{Life-Harness}}

\title{Adapting the Interface, Not the Model: Runtime Harness Adaptation for Deterministic LLM Agents}

\author{
\textbf{Tianshi Xu$^\dagger$} \quad
\textbf{Huifeng Wen$^\dagger$} \quad
\textbf{Meng Li} \\
Peking University \\
\texttt{\{tianshixu, wenhuifeng\}@stu.pku.edu.cn, meng.li@pku.edu.cn}\\
\textsuperscript{$^\dagger$}Equal contribution.
}

\begin{document}
\maketitle

\begin{abstract}
    LLM agents are shaped not only by their language models, but also by the runtime harness that mediates observation, tool use, action execution, feedback interpretation, and trajectory control. While existing agent adaptation methods mainly update model parameters, many failures in deterministic, rule-governed domains stem from mismatches at the model--environment interface.
    We propose \method{}, a lifecycle-aware runtime harness that improves frozen LLM agents without changing model weights or evaluation environments. \method{} evolves from training trajectories by converting recurring interaction failures into reusable interventions across environment contracts, procedural skills, action realization, and trajectory regulation, and remains fixed for evaluation on unseen tasks.
    We evaluate \method{} across eight environments from AgentBench, $\tau$-bench, $\tau^2$-bench, and AppWorld using 18 open-weight backbones, and further test three frontier models on seven environments. With Qwen3-4B-Instruct, \method{} exceeds Meta-Harness and AHE by 28.7 and 7.5 percentage points in mean accuracy across the seven shared environments, respectively, while using 8.5\% and 68.4\% fewer optimization tokens. The evolved harnesses also benefit GPT 5.5, Opus 4.8, and Gemini 3.1 Pro, yielding an average accuracy gain of 3.5 percentage points. Code is available at \href{https://github.com/Tianshi-Xu/Life-Harness}{GitHub}. 
    \begin{figure}[h]
        \centering
        \includegraphics[width=1.0\linewidth]{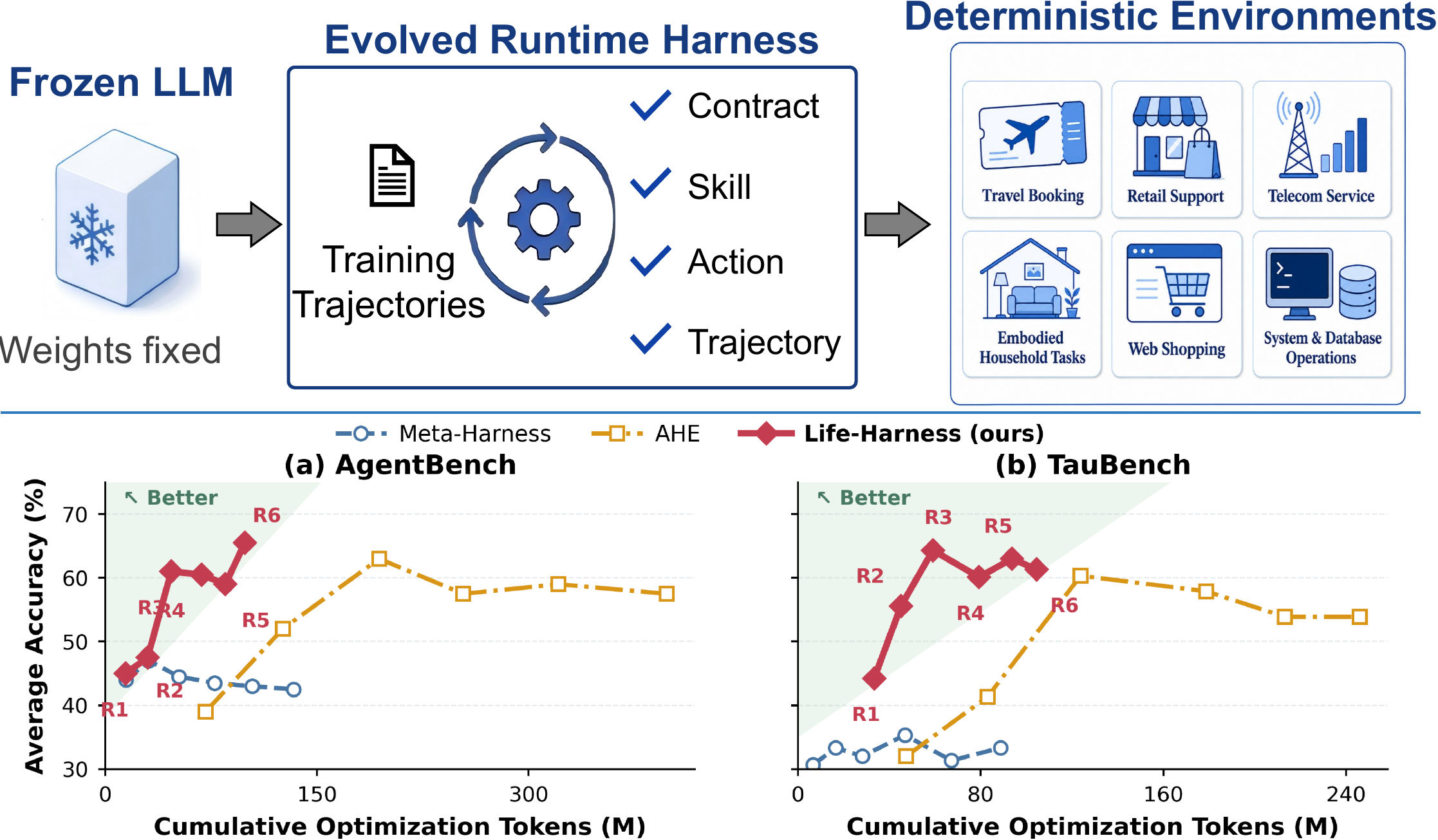}
        \caption{Adapting the runtime harness, not the model. \method{} evolves reusable runtime interventions from training trajectories while keeping LLM weights frozen (top), achieving superior accuracy-efficiency trade-offs over Meta-Harness and AHE across AgentBench and TauBench (bottom).}
        \label{fig:abstract}
    \end{figure}
\end{abstract}
\section{Introduction}
\begin{figure*}[!tb]
    \centering
    \includegraphics[width=0.9\linewidth]{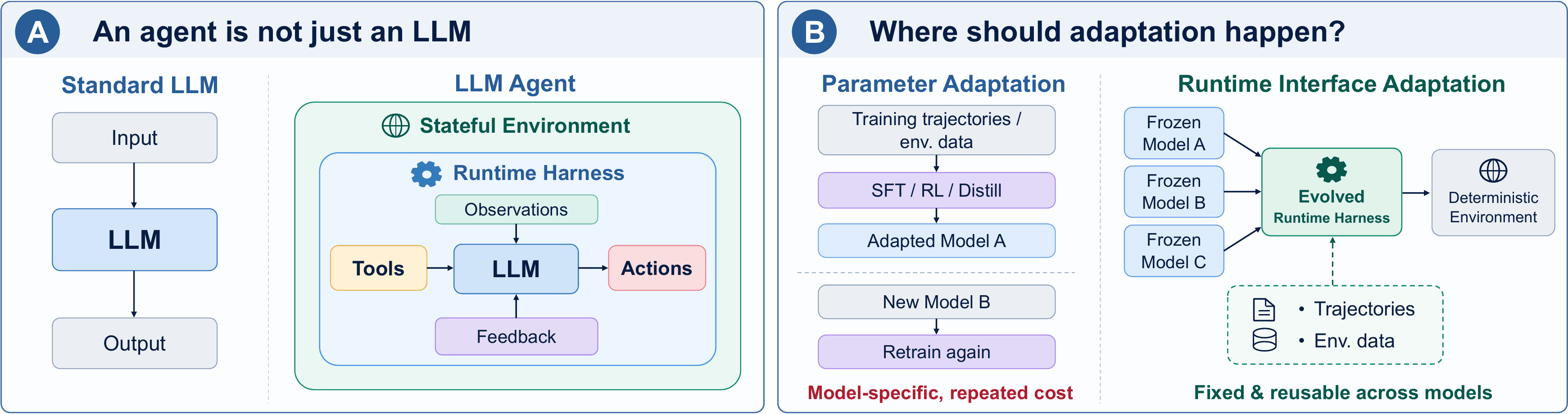}
    \caption{(a) An agent is not just an LLM: its behavior is shaped by the runtime harness that mediates observations, tools, actions, and feedback. 
    (b) We adapt this runtime interface, rather than model parameters, yielding a fixed and reusable harness for frozen agents across deterministic environments.}
    \label{fig:intro}
\end{figure*}
An LLM agent is not just an LLM. As shown in Figure~\ref{fig:intro} (a), it is a model embedded in a stateful interaction loop: the environment exposes observations, the runtime system specifies available tools and actions, the model emits an action or tool call, an executor applies it to the environment, and the resulting feedback updates the next decision~\cite{wang2024executable,anthropic2026claudecode}. The resulting behavior is therefore determined not only by the model, but also by the runtime harness that mediates how the model observes the environment, understands tools, realizes actions, and interprets feedback. This system-level view is increasingly important across agent applications, including software engineering assistants~\cite{yang2024swe,openai2026codexcli,opencode2026}, web navigation~\cite{zhou2024webarena}, database manipulation~\cite{lei2025spider}, operating-system control~\cite{xie2024osworld,liu2024agentbench}, tool-using business workflow agents~\cite{yao2024tau_tau-bench,barres2025tau_tau2-bench}, and embodied interaction~\cite{shridhar2020alfworld}.

Despite this, agent adaptation is commonly framed as model adaptation, including model scaling~\cite{dubey2024llama,qwen2.5}, fine-tuning~\cite{chen2023fireact,prabhakar2026apigen_Salesforce_xLAM2}, reinforcement learning~\cite{guo2025deepseek_Deepseek-r1,qi2025webrl}, and distillation~\cite{kang2026distilling,agarwal2024policy}.
These approaches are powerful but implicitly absorb domain-specific behavior into model parameters. 
In deterministic, rule-governed domains, much of the relevant structure instead lives outside the model: tool schemas, API contracts, admissible actions, feedback rules, stopping conditions, and recovery strategies. 
This distinction is reflected in the gap between static capability and interactive performance. 
For example, Qwen3.5-4B achieves 74.0\% on HMMT~Feb~\cite{hmmt2026}, a competition-level mathematical reasoning benchmark, but only 43.1\% on ALFWorld~\cite{shridhar2020alfworld}, a deterministic embodied interaction benchmark. 
Such failures often arise from model--environment mismatches rather than limited reasoning ability. These include poorly structured observations, misunderstood tool contracts, non-executable actions, feedback that fails to trigger recovery, or degenerate trajectory dynamics.
As illustrated in Figure~\ref{fig:intro} (b), this suggests another route to improvement: exposing stable environment-side structure via the runtime interface where the model observes, acts, and recovers.

Recent work has explored runtime harness optimization for frozen LLM agents, including reasoning-time compute control for mathematical reasoning~\cite{zheng2026llms_AutoTTS_LLMs-Improving-LLMs}, online workspace adaptation in interactive games~\cite{sarafian2026workspace_DREAMTEAM,karten2026continual_Continual-Harness}, harness-flag tuning~\cite{sengupta2026harbor_HARBOR}, and automated harness-code search~\cite{lee2026meta_Meta-harness,lin2026agentic}.
While these approaches show promise for harness optimization, their applicability and efficiency remain limited: some target specific domains or predefined configuration spaces, while broader code-level search can incur substantial optimization costs without explicitly leveraging the structure of recurring interaction failures.
We instead focus on deterministic agent domains, where the harness serves as a stable runtime interface between a frozen model and a rule-governed environment. 
In this setting, recurring failures can be localized to specific stages of the interaction lifecycle, allowing us to evolve failure-specific interface interventions from training trajectories and then fix them during evaluation on unseen tasks. 
This raises our central question: \emph{Can training trajectories reveal stable failure structures that can be converted into a structured runtime interface to improve LLM agents across unseen tasks and diverse model architectures?}

We answer this question with \method{}, a lifecycle-aware runtime harness for deterministic LLM agents. 
Rather than updating model parameters or searching over unconstrained harness code~\cite{lee2026meta_Meta-harness,lin2026agentic}, \method{} adapts the runtime layer that mediates how a frozen model observes the environment, uses tools, realizes actions, interprets feedback, and recovers from degenerate trajectories. 
To model this runtime interface, \method{} organizes adaptation into four lifecycle layers. 
The \textit{Environment Contract Layer} calibrates tool descriptions and interface constraints before interaction. 
The \textit{Procedural Skill Layer} distills reusable procedures from trajectories and retrieves them for the current task and state. 
The \textit{Action Realization Layer} validates and canonicalizes model-generated actions before execution, preventing deterministic failures. 
The \textit{Trajectory Regulation Layer} monitors post-execution dynamics, detects degenerate patterns such as repetition, stagnation, invalid retries, or budget exhaustion, and triggers recovery when needed. 

\method{} evolves through an iterative cycle in which an evolution LLM diagnoses recurring failures in training trajectories and proposes targeted updates to the four runtime layers. Each candidate is validated and compared with the current harness on the same training batch, and retained only if accuracy improves. The selected harness is then frozen for evaluation on unseen tasks.

We evaluate~\method~on eight agent environments from AgentBench~\cite{liu2024agentbench}, $\tau$-bench~\cite{yao2024tau_tau-bench}, $\tau^2$-bench~\cite{barres2025tau_tau2-bench}, and AppWorld~\cite{trivedi2024appworld}. With Qwen3-4B-Instruct~\cite{qwen3technicalreport},~\method~achieves a mean accuracy of {64.45\%} across the seven AgentBench and $\tau$ environments, exceeding Meta-Harness~\cite{lee2026meta_Meta-harness} and AHE~\cite{lin2026agentic} by \textbf{28.7} and \textbf{7.5} percentage points (pp), respectively. It uses \textbf{8.5\%} fewer evolution tokens than Meta-Harness and \textbf{68.4\%} fewer than AHE. The harnesses evolved with Qwen3-4B-Instruct are reused across the other 17 instruction-tuned, reasoning, and agent-specialized backbones, improving \textbf{132 of 144} model--environment settings across all eight environments. On AppWorld, mean accuracy improves by 5.3 pp. The evolved harnesses also benefit frontier models on the seven AgentBench and $\tau$ environments, yielding mean accuracy gains of 2.8, 4.4, and 3.4 pp for GPT 5.5, Opus 4.8, and Gemini 3.1 Pro, respectively.

Our contributions are threefold: 
(1) We formulate \emph{harness-based runtime interface adaptation} for deterministic LLM agents, framing agent improvement as evolving the reusable interface between a frozen model and a rule-governed environment. 
(2) We introduce \textbf{~\method}, a lifecycle-aware framework that converts recurring trajectory failures into targeted interventions across {environment contracts}, {procedural skills}, {action realization}, and {trajectory regulation}. 
(3) We demonstrate {cross-model effectiveness} across eight environments and 18 backbones, and establish higher mean accuracy than Meta-Harness and AHE with lower evolution token budgets. Together, these results demonstrate the effectiveness and reusability of structured {runtime interface adaptation}.
\section{Related Work}

\noindent \textbf{Harness Optimization.} Prior work adapts agent architectures, workflows, and environment interfaces. ADAS~\cite{hu2025automated_Automated-design-of-agentic-systems} and AFlow~\cite{zhang2025aflow_Aflow-Automating-agentic-workflow-generation} optimize architectures and workflows, while WALL-E~\cite{zhou2026wall} learns environment rules for planning. Closer to runtime interaction, ALIGN~\cite{liu2025agent} generates environment interfaces, and AgentSquare~\cite{shang2025agentsquare} searches modular agent architectures. Recent work further optimizes runtime scaffolds around frozen LLMs~\cite{zheng2026llms_AutoTTS_LLMs-Improving-LLMs,sengupta2026harbor_HARBOR,sarafian2026workspace_DREAMTEAM,karten2026continual_Continual-Harness,lou2026autoharness,zhang2026self}. Meta-Harness~\cite{lee2026meta_Meta-harness} and AHE~\cite{lin2026agentic} address automated harness evolution, which are closely related to our work. Meta-Harness searches harness programs using prior candidate code, scores, and execution traces. AHE exposes coding-agent harness components as editable files, organizes execution traces for analysis, and checks whether subsequent evaluation results match the expected effects of proposed changes. Both approaches support automated harness evolution and cross-model transfer. However, organizing optimization around programs or editable components does not by itself specify where an intervention belongs in the interaction loop or what responsibility it should carry. 
Concurrent work HarnessX~\cite{chen2026harnessx} represents harnesses as typed, composable primitives and uses multi-agent analysis of execution traces to evolve their composition. \method{} instead organizes harness adaptation around four explicit roles in the interaction lifecycle: environment contracts before interaction, procedural skills during task conditioning, action realization before execution, and trajectory regulation after feedback. These roles map recurring failures to targeted edits.

\noindent \textbf{Prompt Adaptation Methods.} Prompt optimizers rewrite instructions, demonstrations, or templates using feedback, including PromptWizard, OPRO, TextGrad, GEPA and ACE ~\cite{agarwal2024promptwizard,ORPO_yang2024large_Large-language-models-as-optimizers,yuksekgonul2024textgrad_Textgrad-Automatic-differentiation-via-text,agrawal2025gepa_Gepa-Reflective-prompt-evolution-can-outperform-reinforcement-learning,zhang2026agentic}. \method{} extends adaptation beyond prompts to the broader runtime harness interface.

\noindent \textbf{Model-side Adaptation for LLM Agents.} Instruction tuning, tool-use fine-tuning, reinforcement learning, and distillation improve agents through parameter updates~\cite{qwen2.5,prabhakar2026apigen_Salesforce_xLAM2,guo2025deepseek_Deepseek-r1,xu2026cleaner,qian2026toolrl,lu2025onpolicydistillation}. \method{} complements these approaches by adapting the runtime interface around existing model checkpoints while keeping their parameters fixed.

\section{From Parameter Adaptation to Runtime Interface Adaptation}
\label{sec:preliminaries}

\subsection{LLM Agents as Runtime Systems}
\label{subsec:agents-runtime}

An LLM agent is a policy \(\pi_\theta\) interacting with a stateful environment \(E\), including a user \(U\) when applicable. An episode specifies a task \(x\), an environment contract \(C\), and a step budget \(B\). The contract \(C\) defines tools, argument and feedback formats, answer formats, and task-specific policies. Initialization yields \((s_0,o_0)=E.\textsc{Init}(x)\). 
At step \(t\), the interaction history is
\begin{equation*}
    \tau_t = (C,x,o_0,a_0,o_1,\ldots,a_{t-1},o_t).
\end{equation*} The agent selects an action and receives environment feedback:
\begin{align*}
    a_t &\sim \pi_\theta(\cdot\mid\tau_t),\\
    (s_{t+1},o_{t+1}) &= E.\textsc{Step}(s_t,a_t).
\end{align*}
Actions include tool calls, text commands, and final answers; text actions can be represented as pseudo-tools. The trajectory appends \((a_t,o_{t+1})\), including error feedback, until task completion, environment termination, or budget exhaustion.

\subsection{Parameter vs. Runtime Interface Adaptation}

Given training trajectories \(\mathcal{T}_{\mathrm{train}}\), conventional parameter adaptation updates model weights, whereas runtime interface adaptation updates the harness \(H\) while keeping \(\theta\) fixed:
\begin{align*}
    \theta' &\leftarrow \mathcal{A}_{\mathrm{param}}(\theta,\mathcal{T}_{\mathrm{train}}),\\
    H' &\leftarrow \mathcal{A}_{\mathrm{harness}}(H,\mathcal{T}_{\mathrm{train}}).
\end{align*}
Parameter adaptation embeds task-specific structure in model weights and is \textbf{model- and task-specific}. Our complementary approach adapts how the model interacts with the environment without changing model weights or evaluation protocols. It is \textbf{environment-specific} but \textbf{model-agnostic}: the adapted harness can be reused across model backbones following the same interaction protocol, without retraining.

\section{Method: \method{} }
\label{sec:method}

We propose \method{}, an evolving lifecycle runtime harness for
deterministic LLM agents. The harness adapts the model--environment interface
rather than model weights. It operates on the interaction loop defined in
Section~\ref{subsec:agents-runtime}: the environment contract \(C\), the task
description \(x\), the environment state \(s_t\), the model action \(a_t\), and
the trajectory \(\tau_t\).
\subsection{Failure Diagnosis}
\label{sec:failure-diagnosis}
\begin{figure}[!tb]
    \centering
    \includegraphics[width=1.0\linewidth]{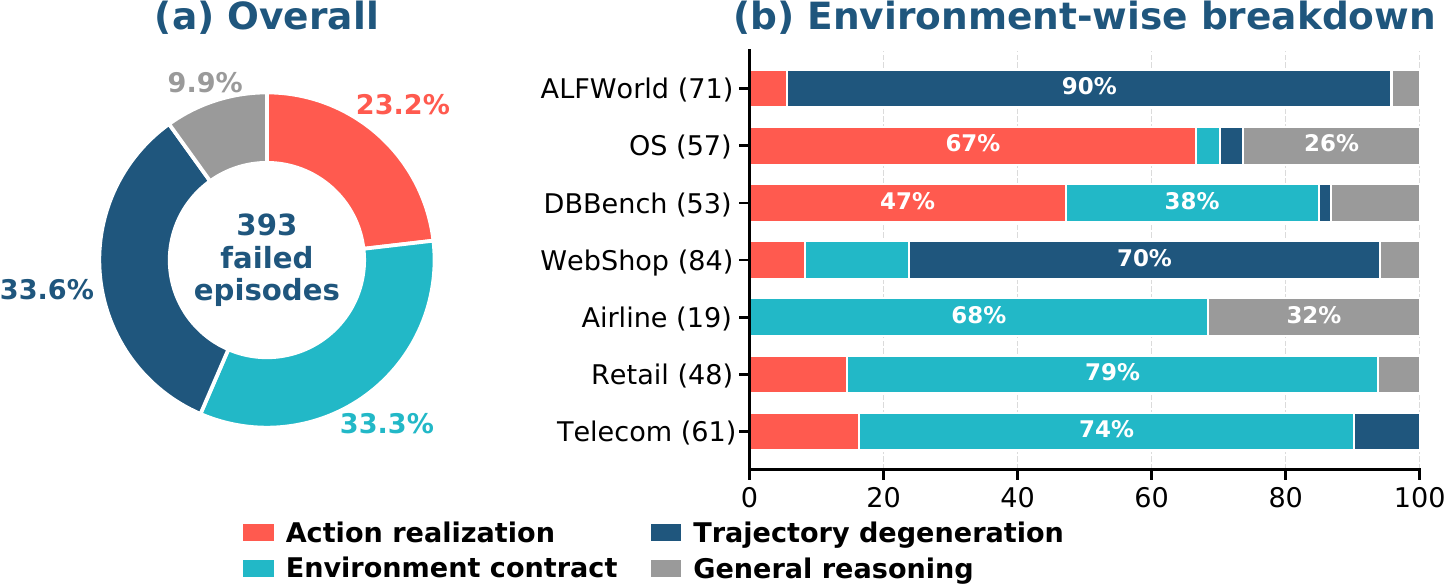}
    \caption{Failure diagnosis on training tasks.}
    \label{fig:failure}
\end{figure}
\begin{figure*}[!tb]
    \centering
    \includegraphics[width=1.0\linewidth]{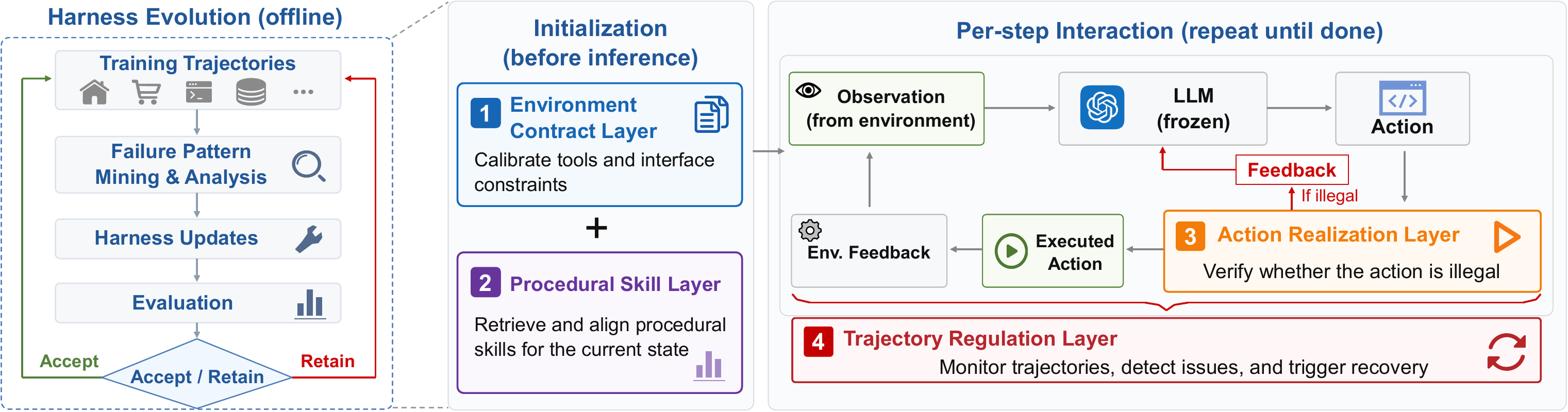}
    \caption{
    Overview of \method{}. The harness adapts the model-environment
    interface through four lifecycle layers spanning before interaction, task
    conditioning, before environment execution, and after execution.
    }
    \label{fig:method}
\end{figure*}
Before designing the harness, we first diagnose the primary failure modes of baseline agents. We evaluate a frozen Qwen3-4B-Instruct on training tasks across diverse interactive agent environments, and manually review LLM-generated failure annotations. For each failed episode, we assign a primary failure type according to the earliest dominant bottleneck in the agent--environment loop, following the classification rules detailed in Appendix~\ref{app:failure-taxonomy}. The resulting taxonomy is summarized in Figure~\ref{fig:failure}. We identify four recurring categories. \textbf{Action realization failures} occur when the model's intent is plausible but not expressed in an environment-executable form, such as free-form actions or missing arguments. \textbf{Environment contract mismatches} occur when an action is syntactically executable but violates the intended tool usage or calling protocol. \textbf{Trajectory degeneration} occurs when individual actions are valid, but the episode falls into repetition, stagnation, or ineffective recovery. The remaining \textbf{general reasoning failures} arise from incorrect inference, computation, or decision-making despite largely following the protocol.

As shown in Figure~\ref{fig:failure}, deterministic agent failures are heterogeneous: all four categories occur in practice, and the dominant mode varies substantially across environments. 
This motivates a harness with multiple intervention points, spanning environment-contract clarification, procedural retrieval, pre-execution action validation, and post-execution trajectory regulation.

\subsection{Overview}
\label{sec:method-overview}

Guided by the failure diagnosis above, \method{} comprises four layers integrated across different stages of the agent lifecycle. Figure~\ref{fig:method} provides an overview of \method{}.

\ding{182} \textbf{Environment Contract Layer} operates before interaction. It makes stable environment constraints explicit, including tool-use rules, policy constraints, and common pitfalls that agents frequently encounter in the target environment.

\ding{183} \textbf{Procedural Skill Layer} operates at the task-conditioning stage. It maintains a skill library distilled from training trajectories and retrieves relevant skills based on the user's task description.

\ding{184} \textbf{Action Realization Layer} operates after the model outputs an action and before the environment executes it. It verifies whether the action is executable under the environment contract, canonicalizes unambiguous interface-level errors, and blocks actions that would deterministically fail. This layer ensures that the model's intended operation is reliably mapped to a valid tool call or environment action.

\ding{185} \textbf{Trajectory Regulation Layer} operates after environment
feedback is returned. It monitors the updated trajectory for non-progressing patterns such as repetition, stagnation, or budget exhaustion, and triggers recovery when needed. This layer specifically targets trajectory degeneration.

Together, these layers adapt the runtime interface through which the model
interacts with the environment. The model weights remain fixed, and the
evaluation environment is unchanged.

\subsection{Detailed Design of \method}

\subsubsection{Environment Contract Layer}
\label{sec:h3}

This layer makes stable environment constraints explicit before interaction by adapting the model-visible contract \(C\). Formally, it produces an enhanced contract $C' = C \oplus \Delta_C$, where \(\Delta_C\) contains concise updates derived from environment policies,
API behavior, and recurring failures in training trajectories. The enhanced contract \(C'\) is shown to the model in place of \(C\), enabling the agent to better utilize the given tools.
In practice, \(\Delta_C\) may specify how tools should be called, which actions are admissible under the environment protocol, and which environment-specific pitfalls should be avoided.

\subsubsection{Procedural Skill Layer}
\label{sec:h5}

This layer provides non-parametric guidance from training
trajectories. A skill is a compact and reusable strategy that captures the essence of how to accomplish specific subtasks.

Let \(\mathcal{S}\) be the skill memory constructed from training trajectories.
For a task description \(x\), the harness retrieves relevant skills:
\[
    \mathcal{K}_x =
    \operatorname{TopK}_{k \in \mathcal{S}}
    \operatorname{score}(x,k),
\]
where \(\operatorname{score}\) is implemented with BM25 in our experiments. The
retrieved skills \(\mathcal{K}_x\) are inserted into the initial system prompt to guide the model on how to solve specific common problems.

\subsubsection{Action Realization Layer}
\label{sec:h2}

This layer operates after the model outputs an action and
before the environment executes it. Given the model action \(a_t\), the current
trajectory \(\tau_t\), and the current state \(s_t\), the
layer either submits this action \( a_t\) to the environment or
returns a model-visible block message \(m_t\):
\begin{align*}
    &z_t =
    \textsc{RealizeAction}(a_t, \tau_t, s_t)\\
    &z_t \in \{\textsc{EXEC}(a_t), \textsc{Block}(m_t)\}.
\end{align*}
It uses deterministic environment evidence, such as tool schemas, admissible action sets, argument constraints, and task policies, to enforce the prevention of erroneous tool calls at the execution level.

\subsubsection{Trajectory Regulation Layer}
\label{sec:h4}

The Trajectory Regulation Layer monitors the interaction after environment
execution. Many agent failures are self-reinforcing: the agent repeats the same invalid command, loops between equivalent states, or exhausts the budget without making progress. Such failures are often detectable from trajectory-level patterns rather than deep semantic understanding.
Given the executed action, returned observation, remaining budget, and
environment evidence, the layer computes:
\[
    r_t =
    \textsc{RegulateTrajectory}(\tau_t,a_t,o_{t+1},b_t)
\]
where \(b_t=B-t-1\) is the remaining budget. The output \(r_t\) may be empty, a soft recovery message, a warning regarding repeated failures, or a stronger corrective directive when the trajectory exhibits clear signs of degradation.

Together, the four layers adapt the runtime interface while preserving model weights and evaluation protocols. Algorithm~\ref{alg:life} summarizes their integration.

\begin{algorithm}[t]
    \caption{\method{} loop}
    \label{alg:life}
    \small
    \KwIn{task \(x\), environment \(E\), contract \(C\), budget \(B\)}
    \(C' \leftarrow \textsc{EnvContract}(C)\), 
    \(x' \leftarrow \textsc{SkillLayer}(x)\)\;
    \(s_0,o_0 \leftarrow E.\textsc{Init}(x')\), 
    \(\tau_0 \leftarrow (C',x',o_0)\)\;
    \For{\(t=0,\ldots,B-1\)}{
        \(a_t \leftarrow \textsc{LLM}(\tau_t)\)\;
        \(z_t \leftarrow \textsc{RealizeAction}(a_t,\tau_t,s_t)\)\;
        \((s_{t+1},o_{t+1}) \leftarrow 
        \textsc{ExecuteOrBlock}(E,s_t,z_t)\)\;
        \(\tau_{t+1} \leftarrow \tau_t \oplus (z_t,o_{t+1})\)\;
        \(\tau_{t+1} \leftarrow 
        \tau_{t+1} \oplus \textsc{RegulateTrajectory}(\tau_t,a_t,o_{t+1},b_t)\)\;
        \If{\textsc{IsEnd}\((s_{t+1},o_{t+1})\)}{\textbf{break}}
    }
\end{algorithm}

\subsection{Trajectory-Driven Harness Evolution}
\label{sec:harness-construction}

\method{} evolves its harness through a repeated cycle of trajectory collection, failure analysis, harness updates, and evaluation, as illustrated in Figure~\ref{fig:method}. Each round starts by running the frozen solver with the current harness on a batch of training tasks. The resulting trajectories show which actions the agent took, what feedback it received, and where progress broke down. An evolution LLM examines these trajectories alongside the current harness to identify recurring failures and propose targeted changes. Throughout evolution, the solver, environment, scoring rules, and interaction budgets remain fixed, and only training evidence is used.

The four layers guide these changes. Misunderstood tool-use rules motivate updates to the Environment Contract Layer; ineffective subtask strategies motivate reusable guidance in the Procedural Skill Layer; malformed or inadmissible actions motivate checks in the Action Realization Layer; and repeated actions or stalled progress motivate recovery interventions in the Trajectory Regulation Layer. The evolution LLM can add, revise, or remove mechanisms within these layers.

Evaluation then determines whether to retain the updated harness in two stages. First, interface checks and replay on training traces test execution validity; failed checks are returned to the evolution LLM for up to three repair attempts. A valid candidate is then evaluated against the current harness on the same training batch using the benchmark's native scorer. It replaces the current harness only if accuracy improves. Otherwise, the current harness is retained, and the next round uses a new training batch to expose additional failure patterns. After evolution, the retained harness is frozen for held-out evaluation. Appendix~\ref{app:evolution-prompts} provides the detailed protocol and prompt template.

\section{Experiments}
\subsection{Experimental Setup}
\label{sec:exp-setup}

\textbf{Benchmarks.}
We evaluate \method{} on four benchmark suites: \(\tau\)-bench~\cite{yao2024tau_tau-bench}, \(\tau^2\)-bench~\cite{barres2025tau_tau2-bench}, AgentBench~\cite{liu2024agentbench}, and AppWorld~\cite{trivedi2024appworld}, covering eight agent environments: {Airline}, {Retail}, {Telecom}, {ALFWorld}~\cite{shridhar2020alfworld}, {WebShop}~\cite{yao2022webshop}, {OS}, {DBBench}, and {AppWorld}. 

\noindent \textbf{Baselines.} We compare \method{} with Meta-Harness~\cite{lee2026meta_Meta-harness}, AHE~\cite{lin2026agentic}, ALIGN~\cite{liu2025agent}, and AgentSquare~\cite{shang2025agentsquare}. For Meta-Harness and AHE, we run harness evolution using open-source implementations under the same conditions as \method{}, including the solver, evolution LLM, and rollout and evaluation settings. Three methods undergo six evolution rounds using identical training batches of up to 50 tasks per round. For ALIGN and AgentSquare, we evaluate their released final harnesses on ALFWorld and WebShop.

\noindent \textbf{Models.}
Throughout harness evolution, we use Qwen3-4B-Instruct~\cite{qwen3technicalreport} as the frozen solver and GPT-5.5 as the evolution LLM.
Held-out test data are excluded from evolution.
The selected harness is frozen and reused to evaluate 17 additional open-weight model backbones, spanning Qwen~\cite{qwen2.5,qwen3technicalreport,qwen3.5}, Llama~\cite{dubey2024llama}, and xLAM~\cite{prabhakar2026apigen_Salesforce_xLAM2} families, including instruction-tuned, reasoning, and agent-specialized models. We also evaluate GPT 5.5, Opus 4.8, and Gemini 3.1 Pro on the seven AgentBench and $\tau$ environments using the same frozen harnesses.

\noindent \textbf{Evaluation Parameters.}
All evaluations use a sampling temperature of \(0.0\). For both \(\tau\) benchmarks, we use DeepSeek-V4-Flash~\cite{deepseekai2026deepseekv4} as the user LLM and evaluate each task three times, reporting single-run success and Pass\(^{3}\), which requires success in all three runs. For AgentBench, we evaluate each task once, following the official implementation. Appendix~\ref{app:exp} provides detailed per-task configurations.
\subsection{Main Results}

\begin{table}[t]
\centering
\begingroup
\fontsize{8}{9.5}\selectfont
\setlength{\tabcolsep}{0.5pt}
\renewcommand{\arraystretch}{1.08}

\newcommand{\benchdash}{\noalign{%
    \vskip 1.6pt
    \hbox to\columnwidth{\color{black}%
        \rule{2.2pt}{0.35pt}%
        \xleaders\hbox{%
            \hskip2.4pt\rule{2.2pt}{0.35pt}%
        }\hfill
        \rule{2.2pt}{0.35pt}%
    }%
    \vskip 1.6pt
}}

\begin{tabular*}{\columnwidth}{
    @{\extracolsep{\fill}}lc
    !{\hskip 2pt\color{black}\vrule width 0.45pt\hskip 2pt}
    ccccccc@{}
}
\toprule
\multirow{3}{*}{\textbf{Task}}
& \multicolumn{1}{c}{\multirow{3}{*}{\textbf{Metric}}}
& \multicolumn{1}{c}{\multirow{3}{*}{\textbf{w/o}}}
& \multicolumn{2}{c}{\textbf{Meta-}}
& \multicolumn{2}{c}{\multirow{2}{*}{\textbf{AHE}}}
& \multicolumn{2}{c}{\textbf{Life-}} \\
& & & \multicolumn{2}{c}{\textbf{Harness}}
& \multicolumn{2}{c}{} & \multicolumn{2}{c}{\textbf{Harness}} \\

\cmidrule(lr){4-5}
\cmidrule(lr){6-7}
\cmidrule(l){8-9}

& \multicolumn{1}{c}{} &
& \textbf{Acc.}$\uparrow$ & \textbf{Cost}$\downarrow$
& \textbf{Acc.}$\uparrow$ & \textbf{Cost}$\downarrow$
& \textbf{Acc.}$\uparrow$ & \textbf{Cost}$\downarrow$ \\
\midrule

\multicolumn{9}{@{}l}{\textit{AgentBench}} \\

\textbf{ALFWorld} & Pass@1 & 16.5
& 35.8 & 3.7
& \textbf{89.0} & 9.6
& 88.1 & \textbf{2.4} \\

\textbf{WebShop} & Pass@1 & 29.5
& 42.5 & \textbf{3.7}
& 39.0 & 10.4
& \textbf{45.5} & 3.8 \\

\textbf{OS} & Pass@1 & 21.5
& 25.7 & 2.7
& 37.5 & 9.4
& \textbf{39.6} & \textbf{2.6} \\

\textbf{DBBench} & Pass@1 & 44.0
& 50.3 & 3.3
& 57.3 & 10.5
& \textbf{65.0} & \textbf{1.1} \\

\midrule
Mean & Pass@1 & 27.9
& 38.6 & 3.3
& 55.7 & 9.9
& \textbf{59.5} & \textbf{2.5} \\

\midrule

\multicolumn{9}{@{}l}{\textit{$\tau$-bench}} \\

\multirow{2}{*}{\textbf{Airline}}
& Pass@1 & 35.0
& 11.7 & \multirow{2}{*}{5.3}
& 35.0 & \multirow{2}{*}{6.4}
& \textbf{60.0} & \multirow{2}{*}{\textbf{3.6}} \\

& Pass$^3$ & 5.0
& 0.0 &
& 20.0 &
& \textbf{50.0} & \\

\benchdash

\multirow{2}{*}{\textbf{Retail}}
& Pass@1 & 48.0
& 33.3 & \multirow{2}{*}{\textbf{1.4}}
& 45.8 & \multirow{2}{*}{9.3}
& \textbf{63.0} & \multirow{2}{*}{2.6} \\

& Pass$^3$ & 25.0
& 15.0 &
& 25.0 &
& \textbf{45.0} & \\

\benchdash

\multicolumn{9}{@{}l}{\textit{$\tau^2$-bench}} \\

\multirow{2}{*}{\textbf{Telecom}}
& Pass@1 & 25.8
& 50.8 & \multirow{2}{*}{\textbf{2.1}}
& \textbf{95.0} & \multirow{2}{*}{8.8}
& 90.0 & \multirow{2}{*}{4.3} \\

& Pass$^3$ & 7.5
& 40.0 &
& \textbf{87.5} &
& 77.5 & \\

\midrule

\multirow{2}{*}{Mean}
& Pass@1 & 36.3
& 31.9 & \multirow{2}{*}{\textbf{3.0}}
& 58.6 & \multirow{2}{*}{8.2}
& \textbf{71.0} & \multirow{2}{*}{3.5} \\

& Pass$^3$ & 12.5
& 18.3 &
& 44.2 &
& \textbf{57.5} & \\

\bottomrule
\end{tabular*}
\endgroup

\caption{
Main results with Qwen3-4B-Instruct.
Acc.: accuracy (\%); Cost: evolution tokens ($10^7$).
Costs span Pass@1 and Pass$^3$ rows where applicable.
All means are unweighted across benchmarks.
Bold denotes the highest accuracy or lowest cost.
}
\label{tab:exp_main}
\end{table}

\begin{table}[t]
    \centering
    \fontsize{7}{8.5}\selectfont
    \setlength{\tabcolsep}{0.8pt}
    \begin{tabular*}{\linewidth}{@{\extracolsep{\fill}}lcccccc@{}}
    \toprule
    \multirow{3}{*}{\textbf{Benchmark}}
    & \multicolumn{2}{c}{\textbf{GPT 5.5}}
    & \multicolumn{2}{c}{\textbf{Opus 4.8}}
    & \multicolumn{2}{c}{\textbf{Gemini 3.1}} \\
    & \multicolumn{2}{c}{(High)}
    & \multicolumn{2}{c}{(High)}
    & \multicolumn{2}{c}{Pro} \\
    \cmidrule(lr){2-3} \cmidrule(lr){4-5} \cmidrule(lr){6-7}
    & w/o & w/ & w/o & w/ & w/o & w/ \\
    \midrule
    ALFWorld & 93.6 & \textbf{95.4} & 97.2 & \textbf{98.1} & 96.3 & 96.3 \\
    DBBench & 67.7 & \textbf{75.0} & 71.7 & \textbf{74.3} & 76.5 & \textbf{78.4} \\
    OS & 47.4 & \textbf{53.1} & 50.4 & \textbf{50.8} & 50.0 & 48.6 \\
    Webshop & 75.1 & \textbf{75.8} & 73.8 & \textbf{82.7} & 67.3 & \textbf{68.1} \\
    \midrule
    Airline & 78.3 & \textbf{80.0} & 80.0 & \textbf{82.8} & 75.9 & \textbf{81.5} \\
    Retail & 72.5 & 72.5 & 72.5 & \textbf{84.2} & 68.3 & \textbf{80.8} \\
    \midrule
    Telecom & 97.3 & \textbf{99.5} & 93.4 & \textbf{96.8} & 93.2 & \textbf{97.7} \\
    \bottomrule
    \end{tabular*}
    \caption{Frontier-model results (\%) without (w/o) and with (w/) Life-Harness.}
    \label{tab:frontier_models}
\end{table}
\begin{figure}[!tb]
    \centering
    \includegraphics[width=1.0\linewidth]{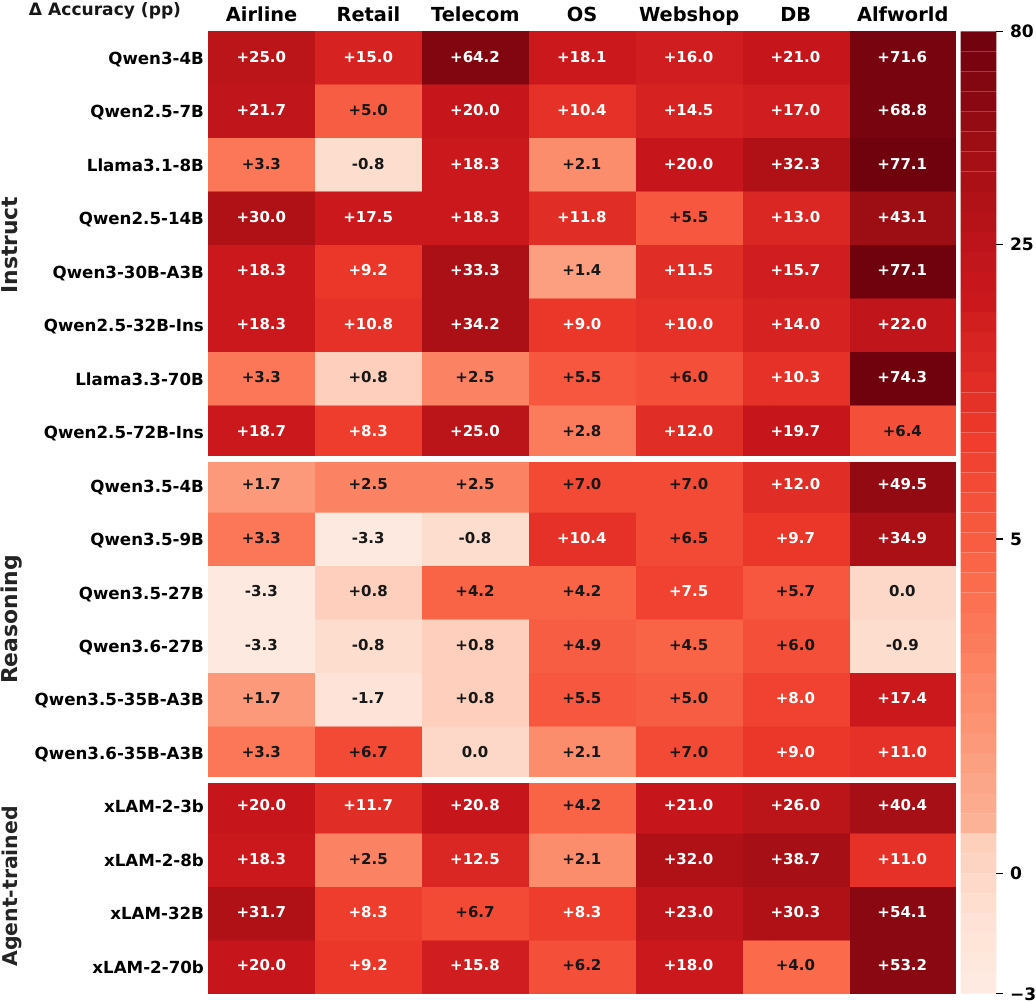}
    \caption{Absolute performance improvement across 18 model backbones and 7 benchmarks.}
    \label{fig:exp_heatmap}
\end{figure}
\noindent \textbf{Accuracy and Evolution Efficiency.} Table~\ref{tab:exp_main} compares \method{} with Meta-Harness and AHE using Qwen3-4B-Instruct. \method{} improves mean Pass@1 accuracy from 27.9\% to 59.5\% on AgentBench and from 36.3\% to 71.0\% across the two $\tau$ suites, achieving the highest accuracy on five of seven environments. Compared with AHE, it achieves higher mean accuracy on both groups while requiring fewer evolution tokens in every environment. AHE performs better on ALFWorld and Telecom, whereas \method{} provides a more favorable overall accuracy--cost trade-off. Across all seven environments, \method{} exceeds Meta-Harness and AHE by 28.7 and 7.5 percentage points in mean accuracy, respectively, while using 8.5\% fewer evolution tokens than Meta-Harness and 68.4\% fewer than AHE. Figure~\ref{fig:exp_heatmap} shows the full accuracy gains across 18 backbones and seven environments, with improvements in 116 of 126 model--environment settings; complete scores appear in Appendix Table~\ref{tab:appendix-full-results}.
\begin{figure}[!tb]
    \centering
    \includegraphics[width=1.0\linewidth]{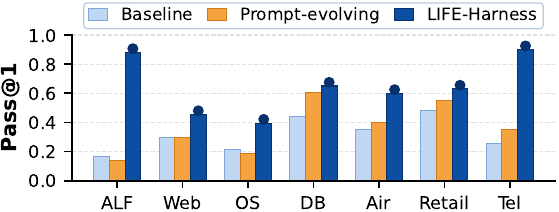}
    \caption{Comparison with prompt evolving method.}
    \label{fig:exp_prompt}
\end{figure}
\begin{table}[t]
    \centering
    \fontsize{7}{8.5}\selectfont
    \setlength{\tabcolsep}{2pt}
    \begin{tabular*}{\linewidth}{@{\extracolsep{\fill}}lccc@{}}
    \toprule
    \textbf{Setting} & \textbf{w/o} & \textbf{Life-Harness} & \textbf{Improved} \\
    \midrule
    Mean (18 backbones) & 25.50 & \textbf{30.76} & 16/18 \\
    \bottomrule
    \end{tabular*}
    \caption{AppWorld mean Pass@1 accuracy (\%) across 18 backbones. Improved counts backbones with strictly higher accuracy under Life-Harness.}
    \label{tab:appworld}
\end{table}
\begin{table}[!tb]
    \centering
    \huge
    \setlength{\tabcolsep}{4.5pt}
    \resizebox{1.0\linewidth}{!}{
    \begin{tabular}{lccc|cccc}
    \toprule
    \multirow{2}{*}{\textbf{Setting}}
    & \multicolumn{2}{c}{\textbf{$\tau$-bench}}
    & $\tau^2$-bench
    & \multicolumn{4}{c}{\textbf{AgentBench}} \\
    \cmidrule(lr){2-3} \cmidrule(lr){4-4} \cmidrule(lr){5-8}
    & \textbf{Airline} & \textbf{Retail} & \textbf{Telecom} & \textbf{ALFWorld} & \textbf{WebShop} & \textbf{OS} & \textbf{DBBench} \\
    \midrule
    \method
        & 0.0\% & 0.0\% & 0.0\% & 0.0\% & 0.0\% & 0.0\% & 0.0\% \\
    w/o Contract & -8.3\% & -17.5\% & -16.0\% & -1.0\% & -4.4\% & -14.1\% & -16.9\% \\
    w/o Skill & -8.3\% & -15.9\% & -17.4\% & -1.0\% & -2.2\% & -14.1\% & -3.1\% \\
    w/o Action & -61.7\% & -15.9\% & -10.1\% & -1.0\% & -6.6\% & -59.6\% & -4.6\% \\
    w/o Trajectory & -3.3\% & -16.7\% & -36.2\% & -86.5\% & -26.4\% & -14.1\% & -4.6\% \\
    \bottomrule
    \end{tabular}
    }
    \caption{
    Leave-one-layer-out ablation on Qwen3-4B-Instruct. Values report the relative \textbf{accuracy drop} compared with the full \method. ``Contract'', ``Skill'', ``Action'', and ``Trajectory'' denote the Environment Contract, Procedural Skill, Action Realization and Trajectory Regulation layers, respectively.
    }
    \label{tab:ablation}
\end{table}
\begin{figure}[!t]
    \centering
    \includegraphics[width=1.0\linewidth]{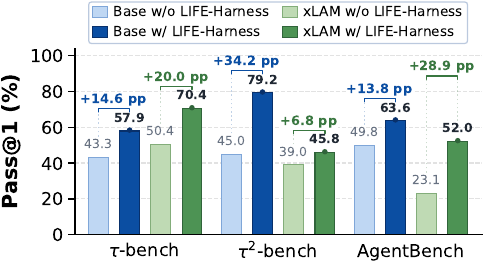}
    \caption{Runtime harnessing improves both Qwen2.5-32B-Instruct (Base) and its tool-trained derivative, xLAM-2-32B, across all three benchmark groups. Bars show Pass@1 accuracy; brackets indicate gains from \method{} in pp.}
    \label{fig:exp_ablation2}
\end{figure}

\noindent \textbf{Evolution Dynamics.} Figure~\ref{fig:abstract} tracks accuracy gains against cumulative evolution tokens over six rounds, showing that \method{} achieves substantial improvements with a modest token budget. Table~\ref{tab:appendix-evolution-data} in the appendix provides complete accuracy and cost records for each environment and round.

\noindent \textbf{Comparison with Prompt Optimization.} Figure~\ref{fig:exp_prompt} compares \method{} with prompt-only optimization~\cite{agrawal2025gepa_Gepa-Reflective-prompt-evolution-can-outperform-reinforcement-learning,ORPO_yang2024large_Large-language-models-as-optimizers} on Qwen3-4B-Instruct. While optimizing the prompt improves accuracy over the baseline, \method{} achieves higher Pass@1 accuracy across all seven environments. These results support adapting the broader runtime interface, including action execution, feedback handling, and trajectory control, beyond the input prompt alone.

\noindent \textbf{Transfer to Frontier Models.} Table~\ref{tab:frontier_models} evaluates the evolved harnesses with GPT 5.5, Opus 4.8, and Gemini 3.1 Pro. Averaged across the seven environments, \method{} improves accuracy by 2.8, 4.4, and 3.4 pp, respectively, with gains in 18 of 21 model--environment settings. These results show that runtime harness adaptation remains beneficial even for highly capable models.

\noindent \textbf{Comparison with ALIGN and AgentSquare.} We evaluate the released open-source harnesses of ALIGN~\cite{liu2025agent} and AgentSquare~\cite{shang2025agentsquare} on ALFWorld and WebShop across 18 backbones. \method{} outperforms both methods across all backbones on ALFWorld and in most settings on WebShop. Complete results and evaluation configurations are provided in Appendix Table~\ref{tab:align-agentsquare}.

\noindent \textbf{Extension to Application Operation.} We further evaluate \method{} on AppWorld~\cite{trivedi2024appworld}, an API-based benchmark for multi-application operation. As shown in Table~\ref{tab:appworld}, \method{} improves mean Pass@1 accuracy from 25.5\% to 30.8\% across 18 backbones, with improvements on 16. Complete results are provided in Appendix Table~\ref{tab:appworld-full-results}.

\subsection{Ablation Study}
\noindent \textbf{All Lifecycle Layers Matter}. Table~\ref{tab:ablation} reports leave-one-layer-out ablations. Removing any harness layer causes substantial accuracy drops on some benchmarks. The impact varies across environments, reflecting their different interaction requirements.

\noindent \textbf{Case Study: Harnessing a Tool-Trained Model.} Figure~\ref{fig:exp_ablation2} compares xLAM-2-32B with its base model, Qwen2.5-32B-Instruct, before and after applying \method{}. xLAM-2-32B undergoes specialized tool-use training on scenarios related to $\tau$-bench~\cite{prabhakar2026apigen_Salesforce_xLAM2}. Adding \method{} improves xLAM-2-32B across all evaluated benchmark groups, showing that runtime adaptation provides additional benefits even after tool-use training. On $\tau$-bench, Qwen2.5-32B-Instruct equipped with \method{} also exceeds the original xLAM-2-32B by 7.5 pp. Together, these results illustrate how adapting the runtime interface can complement capabilities acquired through model training.

\section{Conclusion}
We present \method{}, a lifecycle-aware framework that evolves reusable runtime interventions from agent trajectories across environment contracts, procedural skills, action realization, and trajectory regulation. With model weights and evaluation environments fixed, \method{} improves accuracy across diverse open-weight and frontier models. Comparisons with Meta-Harness and AHE show higher mean accuracy with lower evolution token budgets. These findings support structured runtime interface adaptation as an effective and reusable complement to model training for improving LLM agents.

\section*{Limitations}
Our evaluation covers text- and API-based agent environments. We leave the evaluation of \method{} on visual GUI tasks, such as desktop and mobile application operation, to future work. These settings would extend the empirical scope of runtime harness adaptation to agents that interact through visual observations and graphical actions.

\bibliography{custom}

\appendix
\section{Detailed Harness}
\subsection{Failure Annotation Protocol}
\label{app:failure-taxonomy}

We use an LLM to annotate failed trajectories, and human reviewers check all annotations. For each failed episode, the LLM reads the complete interaction trace and assigns one primary failure category according to the rule-based protocol below. The classification is performed trajectory by trajectory, rather than from aggregate statistics. We use a priority-based protocol: the annotator first checks for action realization failures, then environment contract mismatches, then trajectory degeneration, and finally assigns the episode to the residual category. This priority order prevents later symptoms from hiding earlier interface failures. For example, if an agent writes a tool call in plain text, the environment never executes it, and the episode eventually exhausts the step budget, we annotate it as an action realization failure rather than trajectory degeneration.

\paragraph{Action realization failures.}
This category corresponds to failures targeted by the \emph{Action Realization Layer}. A failed episode is assigned to this category if any of the following conditions hold:
\begin{itemize}
    \item The assistant does not produce executable \texttt{tool\_calls}, but instead writes in \texttt{content} an action or answer that should have been submitted through a tool, such as \texttt{take\_action(\{...\})}, \texttt{answer\_action(\{...\})}, a natural-language command, SQL text, or the final answer text.
    \item The assistant produces a tool call, but the call cannot be executed because of an invalid function name, JSON parsing failure, missing required arguments, or incorrect argument types.
    \item In DBBench, the SQL query is not executable due to interface, dialect, or formatting violations, such as unquoted column names containing spaces, malformed table names, or invalid concatenation syntax.
\end{itemize}
The key criterion is that the model's intent may be reasonable, but it is not submitted in a form executable by the environment.

\paragraph{Environment contract mismatches.}
This category corresponds to failures targeted by the \emph{Environment Contract Layer}. It is considered only when the action interface is executable. A failed episode is assigned to this category if any of the following conditions hold:
\begin{itemize}
    \item The agent uses the wrong tool for a critical step, such as calling a finish tool when an answer-submission tool is required.
    \item The tool choice or calling order violates the task protocol, such as submitting prematurely, skipping a required intermediate tool, or using a search tool to encode constraints that should not be searched directly.
    \item The argument has an incorrect semantic format, although it satisfies the JSON or schema-level interface. For example, in DBBench, the final answer may be required to be an exact value or list, but the agent submits a natural-language explanation, or multi-column results are concatenated into free-form text instead of the required format.
\end{itemize}
The key criterion is that the function call is structurally present and executable, but the agent misunderstands the tool's purpose, boundary, calling protocol, or argument semantics.

\paragraph{Trajectory degeneration.}
This category corresponds to failures targeted by the \emph{Trajectory Regulation Layer}. It is considered only when neither action realization nor environment contract mismatch is the primary cause. A failed episode is assigned to this category if any of the following conditions hold:
\begin{itemize}
    \item The episode ends due to \texttt{task limit reached} or an equivalent budget-exhaustion signal, and the trajectory contains a clear repetition pattern, such as repeatedly issuing the same action, oscillating between two states, repeatedly calling \texttt{look}, \texttt{inventory}, or \texttt{examine}, repeatedly searching or clicking, or repeatedly returning to the same page.
    \item The agent commits to an incorrect strategy early and reinforces it throughout the trajectory, such as searching once and purchasing without comparing candidates or required attributes, or repeatedly visiting already explored locations after failing to find the target.
    \item The environment repeatedly returns no-progress feedback, such as \texttt{Nothing happens} or an unchanged page/state over many steps.
\end{itemize}
The key criterion is that individual actions are usually executable, but the long-horizon interaction fails to use environment feedback to revise the strategy.

\paragraph{Residual reasoning failures.}
A failed episode is assigned to this category when none of the above categories apply. This category includes cases where the tool-use protocol is largely followed, but the agent makes an incorrect reasoning, computation, SQL, retrieval, or value-selection decision. It also includes cases where the agent retrieves the wrong object or answer for reasons not attributable to tool-protocol misunderstanding or trajectory loops, or where the correct task path is attempted but the final value, condition, or filtering logic is wrong.

\subsection{Harness Evolution Protocol and Prompt}
\label{app:evolution-prompts}

The two templates below summarize the core prompts used for harness evolution and validation repair in Section~\ref{sec:harness-construction}. The evolution prompt combines fixed instructions with round-specific inputs; the repair prompt supplies validation feedback within the same proposal. Placeholders denote inputs supplied by the host. These templates omit auxiliary context-management instructions; exact prompts and configurations are recorded with each run's archived source version.

\begin{tcolorbox}[
    colback=gray!3,
    colframe=gray!45,
    title={Harness Evolution Prompt},
    fonttitle=\bfseries,
    breakable,
    enhanced,
    boxrule=0.6pt,
    arc=2pt,
    left=5pt,
    right=5pt,
    top=5pt,
    bottom=5pt
]
\textbf{System instructions.} You are the harness evolution LLM. Improve training-task accuracy by updating the runtime harness. Keep the solver, environment, tasks, native tool schemas, scorer, and interaction budgets fixed. Use only exposed training evidence and permitted tools; never access test trajectories or hidden answers.

Read the current harness, interface contracts, training results, and relevant trajectories. Inspect recurring failures and successful cases to locate where behavior diverges from the task requirements. Distinguish execution infrastructure errors from agent behavior, and treat symptoms as evidence to investigate rather than established causes. Consult previous candidates and their gains or regressions when relevant.

Choose an intervention within the four interfaces: clarify tool-use rules through the Environment Contract Layer; provide reusable strategies through the Procedural Skill Layer; validate, repair, or block actions through the Action Realization Layer; or detect stalled progress and guide recovery through the Trajectory Regulation Layer. Add, revise, or remove mechanisms as justified by the evidence. Edits may span multiple layers, but need not change every layer. Preserve useful behavior, consider regression risks, and keep runtime state local to each episode.

Write the candidate as \texttt{bundle.json} and its referenced artifacts, and run the candidate validation tool. Repair reported defects before handing off a valid candidate. Passing trace replay establishes execution validity, not improved task accuracy. The host evaluates the candidate and current harness on the same complete training batch and accepts only a strict accuracy improvement.

\textbf{Round input.} Optimize the current harness using the following workspace inputs: \texttt{\{current\_harness\}}, \texttt{\{interface\_contracts\}}, \texttt{\{training\_results\}}, \texttt{\{training\_trajectories\}}, and \texttt{\{proposal\_history\}}. Inspect relevant evidence on demand, implement a candidate update, and submit it for evaluation after validation.
\end{tcolorbox}

\begin{tcolorbox}[
    colback=gray!3,
    colframe=gray!45,
    title={Validation Repair Prompt},
    fonttitle=\bfseries,
    breakable,
    enhanced,
    boxrule=0.6pt,
    arc=2pt,
    left=5pt,
    right=5pt,
    top=5pt,
    bottom=5pt
]
Candidate validation failed. This is repair attempt \texttt{\{attempt\}} of 3 for the same proposal.

\textbf{Validation feedback:} \texttt{\{validation\_errors\}}.

Inspect the reported defects, revise the candidate, and rerun validation. Keep the original optimization boundaries unchanged. This repair attempt does not impose a limit on evidence reads or interaction turns. Successful validation permits task-accuracy evaluation; it does not establish an accuracy gain.
\end{tcolorbox}

\subsection{Final Evolved Harness Inventory}
\label{app:harness-inventory}

This appendix summarizes the concrete harness components used in the seven evaluated scenarios. The inventory follows the four lifecycle layers described in the main method section. Detailed implementations of these components are provided in our submitted codebase: \href{https://github.com/Tianshi-Xu/Life-Harness}{GitHub}. The $\tau$-bench and $\tau^2$-bench harness components are listed in Table~\ref{tab:taubench-inventory}, and the AgentBench components are listed in Table~\ref{tab:agentbench-inventory} and Table~\ref{tab:agentbench-inventory-os-db}.

\begin{table*}[t]
    \centering
    \huge
    \setlength{\tabcolsep}{4pt}
    \resizebox{1.0\linewidth}{!}{
    \begin{tabular}{c|c|c|c|c|c}
    \toprule
    \textbf{Task} & \textbf{Raw Train Pool} & \textbf{Train Used for Harness Evolution} 
    & \textbf{Test / Eval} & \textbf{Raw Pool Total} \\
    \midrule
    ALFWorld       & 3,150 & 100 & 109 & 3,259 \\
    DBBench        & 4,803 & 100 & 300 & 5,103 \\
    OS Interaction & 1,000 & 100 & 144 & 1,144 \\
    WebShop        & 9,000 & 100 & 200 & 9,200 \\
    Airline        & 30    & 30  & 20  & 50    \\
    Retail         & 74    & 74  & 40  & 114   \\
    Telecom        & 74    & 74  & 40  & 114   \\
    AppWorld       & 90    & 90  & 57 & 147 \\
    \bottomrule
    \end{tabular}
    }
    \caption{
    Dataset statistics for harness evolution and evaluation.
    }
    \label{tab:data_statistics}
\end{table*}

\begin{table}[t]
    \centering
    \caption{Evaluation sampling and interaction-budget settings.}
    \label{tab:eval-config}
    \resizebox{1.0\linewidth}{!}{
    \begin{tabular}{lcccc}
    \toprule
    Benchmark & Temperature & Context limit & Max tokens per step & Max step \\
    \midrule
    ALFWorld & $0.0$ & 32,768 & 4096 & 50 \\
    DBBench & $0.0$ & 32,768 & 4096 & 15 \\
    OS & $0.0$ & 32,768 & 4096 & 16 \\
    WebShop & $0.0$ & 32,768 & 4096 & 20 \\
    Airline & $0.0$ & 32,768 & 2048 & 200 \\
    Retail & $0.0$ & 32,768 & 2048 & 200 \\
    Telecom & $0.0$ & 32,768 & 2048 & 200 \\
    AppWorld & $0.0$ & 40,960 & 4096 & 50 \\
    \bottomrule
    \end{tabular}
    }
\end{table}

\begin{table*}[t]
\centering
\huge
\scriptsize
\setlength{\tabcolsep}{2.2pt}
\renewcommand{\arraystretch}{0.88}
\begin{tabular}{>{\centering\arraybackslash}m{0.1\textwidth}>{\centering\arraybackslash}m{0.2\textwidth}p{0.7\textwidth}}
\toprule
Domain & Layer & Implemented harness content \\
\midrule
\multirow{26}{*}{\centering Airline}
& \multirow{6}{*}{Environment Contract Layer}
& search tools remind that flight search takes only origin, destination, and date;
search results must be compared by requested cabin price, total travel time, and
seat availability; cancellation tool lists the four valid cancellation conditions
and transfers flying/landed flights to humans; update tool requires complete
round-trip itineraries and reservation-wide cabin changes; booking tool states
payment-method caps, five-passenger cap, bookable-flight status, insurance, and
baggage allowance rules; certificate tool states eligibility, amount cap, and
compensation timing; transfer tool lists the cases that are actually out of
scope. \\
\cmidrule(lr){2-3}
& \multirow{5}{*}{Procedural Skill Layer}
& communicate exact charge/refund after every cancel, update, or booking write;
when multiple reservations exist, inspect candidate reservation details before
acting; use at most one certificate per reservation and split only when separate
bookings still satisfy the user request; submit all outbound and return legs in
one round-trip update; check booking time, airline cancellation, business cabin,
and insurance before cancellation; calculate total card charge or budget impact
before writing; scan all reservations for duplicate, same-day, or schedule-mixup
requests. \\
\cmidrule(lr){2-3}
& \multirow{8}{*}{Action Realization Layer}
& \texttt{book\_reservation} checks duplicate payment IDs, at most one certificate, one
credit card, and three gift cards; \texttt{book\_reservation} checks max five passengers,
connected itinerary, available flight status, enough seats, correct baggage fee,
and exact payment total; \texttt{cancel\_reservation} checks flown-flight guard and the
four cancellation conditions; \texttt{update\_reservation\_flights} blocks basic-economy
flight changes, origin/destination/trip-type changes, missing return legs,
post-departure cabin changes, certificates as payment, and insufficient gift-card
capacity; \texttt{update\_reservation\_baggages} blocks baggage decreases and wrong
nonfree-bag count; \texttt{update\_reservation\_passengers} blocks passenger-count changes;
\texttt{send\_certificate} checks user eligibility, real delayed/cancelled flight status,
prior change/cancel for delay compensation, and amount multiple/cap. \\
\cmidrule(lr){2-3}
& \multirow{5}{*}{Trajectory Regulation Layer}
& \texttt{get\_user\_details} appends one-line summaries for every reservation when a user
has several; \texttt{get\_reservation\_details} appends total paid by payment method,
membership, insurance, cabin, passenger count, free-bag allowance, and current
round-trip legs; \texttt{cancel\_reservation} appends exact refund amount and destination
payment instrument; \texttt{update\_reservation\_flights} appends exact extra charge or
refund; \texttt{book\_reservation} appends total charged by payment instrument. \\
\midrule
\multirow{30}{*}{\centering Retail}
& \multirow{8}{*}{Environment Contract Layer}
& \texttt{cancel\_pending\_order} description says the tool cancels the whole pending order
and accepts only \texttt{no longer needed} or \texttt{ordered by mistake};
\texttt{modify\_pending\_order\_items} description says item modification is once per pending
order and replacement IDs must be same-product, different, and available;
\texttt{modify\_pending\_order\_payment} description requires a saved payment ID and enough
gift-card balance for the full positive total; address tools distinguish profile
address from order shipping address; exchange/return descriptions say one return
or exchange per delivered order and exact item IDs from order details; return
description clarifies \texttt{available=false} does not block returning an owned
item; transfer description lists only out-of-scope cases. \\
\cmidrule(lr){2-3}
& \multirow{6}{*}{Procedural Skill Layer}
& count every variant with \texttt{available=true} for availability-count tasks;
copy hidden target addresses from profile or previous orders before writing;
for pending orders, do not use full-order cancellation when the user rejects that
fallback; for bulk requests, inspect all candidate orders and group writes by
order; exchange all requested items from the same delivered order in one call;
compare numeric option values such as storage, zoom, size, or piece count; keep
theme-scoped requests limited to matching items; use item modification, not a new
order, for pending same-product replacements; after clear user confirmation,
call the write tool instead of summarizing again. \\
\cmidrule(lr){2-3}
& \multirow{7}{*}{Action Realization Layer}
& \texttt{cancel\_pending\_order} checks order status and reason; \texttt{modify\_pending\_order\_items}
checks status, once-only modification, non-empty \texttt{item\_ids}, non-empty
\texttt{new\_item\_ids}, equal list lengths, old item IDs in the order, duplicate
IDs against order quantity, replacement IDs in the catalog, same-product
replacement, and not replacing with the same item; \texttt{modify\_pending\_order\_payment}
checks new method differs and gift card covers the full total; \texttt{exchange\_delivered\_order\_items}
checks delivered status, one exchange/return only, valid old/new IDs, duplicate
old IDs, same-product replacement, and not same item; \texttt{return\_delivered\_order\_items}
checks delivered status, valid return items, duplicate return IDs, and refund to
original method or user's gift card. \\
\cmidrule(lr){2-3}
& \multirow{7}{*}{Trajectory Regulation Layer}
& \texttt{get\_order\_details} appends status-specific routing, in-progress return/exchange
or already-modified warnings, tracking numbers, and order total; \texttt{get\_product\_details}
appends available variant count, cheapest/most-expensive available variants, and
numeric option min/max summaries; \texttt{get\_item\_details} explains that out-of-stock
only blocks new purchases, not returns/exchanges of owned items; exchange writes
append price difference; return writes append refund amount and remaining-item
total when relevant; cancel writes append refund amount; gift-card write paths
can append updated gift-card balance; completion annotators mention other
eligible orders only as conditional follow-up, not as permission to act. \\
\midrule
\multirow{28}{*}{\centering Telecom}
& \multirow{6}{*}{Environment Contract Layer}
& \texttt{send\_payment\_request} description says use it only for overdue bills and only
one awaiting-payment bill at a time; \texttt{resume\_line} description says all overdue
bills must be paid and expired contracts require transfer; \texttt{refuel\_data}
description says line must be active, max 2 GB per call, and price/consent must
be confirmed; \texttt{suspend\_line} description says line must be active and the user
must accept the monthly fee; \texttt{enable\_roaming} description separates account-side
enablement from the user's device-side Data Roaming toggle; transfer description
requires an actual tool call for expired contracts, SIM PIN/in-store support, or
issues outside the tool set. \\
\cmidrule(lr){2-3}
& \multirow{7}{*}{Procedural Skill Layer}
& data issues retrieve a quota-check skill that calls \texttt{get\_data\_usage} before
chasing APN/VPN causes; abroad issues retrieve roaming skills that call
\texttt{enable\_roaming} when needed and always ask the user to toggle Data Roaming;
service-loss issues retrieve the overdue-bill workflow: get bills, send payment
request, wait for payment, resume line, reboot; MMS skills require WiFi Calling
off, mobile data/APN/SIM/app permissions, and quota check before transfer;
multiple-cause skills force the agent to continue after one fix if the symptom
persists; line-selection skill uses the line matching the caller phone number;
SIM PIN skill separates human transfer for PIN lock from fixable billing work;
hard-persona skill switches to one short instruction per turn. \\
\cmidrule(lr){2-3}
& \multirow{6}{*}{Action Realization Layer}
& \texttt{get\_customer\_by\_phone} checks NXX-NXX-XXXX phone format; write tools check
\texttt{customer\_id} format/existence and line ownership; \texttt{send\_payment\_request} checks
bill ownership, overdue status, no expired-contract restoration trap, and no
other awaiting-payment bill; \texttt{resume\_line} checks line is suspended, all overdue
bills are paid, and contract is not expired; \texttt{refuel\_data} checks max 2 GB and
active line status; \texttt{suspend\_line} checks active line status; \texttt{enable\_roaming}
blocks when roaming is already enabled and redirects to device toggle;
\texttt{disable\_roaming} blocks when roaming is already disabled. \\
\cmidrule(lr){2-3}
& \multirow{6}{*}{Trajectory Regulation Layer}
& \texttt{get\_customer\_by\_phone} appends the \texttt{line\_id} matching the caller's phone when the
customer has multiple lines; \texttt{get\_data\_usage} appends quota-exhausted warning and
refuel suggestion; \texttt{get\_details\_by\_id} appends roaming-off, quota-exhausted,
suspended-line, or expired-contract notes; \texttt{get\_bills\_for\_customer} appends overdue
bill IDs and the send-payment-request to resume-line workflow; \texttt{send\_payment\_request}
appends the required customer payment, verification, resume, and reboot steps;
\texttt{enable\_roaming} appends whether the account was newly enabled or already enabled
and reminds the device-side Data Roaming toggle. \\
\bottomrule
\end{tabular}
\caption{$\tau$-bench and $\tau^2$-bench harness inventory under the four lifecycle layers.}
\label{tab:taubench-inventory}
\end{table*}

\begin{table*}[t]
\huge
\centering
\scriptsize
\setlength{\tabcolsep}{2.2pt}
\renewcommand{\arraystretch}{0.86}
\begin{tabular}{>{\centering\arraybackslash}m{0.1\textwidth}>{\centering\arraybackslash}m{0.2\textwidth}p{0.7\textwidth}}
\toprule
Domain & Layer & Implemented harness content \\
\midrule
\multirow{21}{*}{\centering ALFWorld}
& \multirow{4}{*}{Environment Contract Layer}
& task parsing extracts task type, target object, destination receptacle,
required transform, object count, and subgoal chain; task-order hints spell out the
step order for pick/place, clean, heat, cool, look-at, and pick-two tasks inside
\texttt{take\_action}; action-tool schema patching appends the hint so the model sees admissible-action and task-order constraints
on every call. \\
\cmidrule(lr){2-3}
& \multirow{4}{*}{Procedural Skill Layer}
& task-type skill retrieval filters the ALFWorld skill library and BM25-ranks
matches; injectable contract skills tell the agent to pick up the object before
cleaning/heating/cooling and to use \texttt{examine X with desklamp}, not plain
\texttt{examine X}; non-injected library entries document pickup-then-deliver,
two-object staging, unseen-location search, appliance-as-midpoint, ``nothing
happens'' recovery, second-object traps, and loop breaking. \\
\cmidrule(lr){2-3}
& \multirow{4}{*}{Action Realization Layer}
& \texttt{pre\_validate\_action} compares the raw action with the current admissible list;
\texttt{\_gate\_action} canonicalizes close string matches only when the verb is compatible;
invalid non-navigation actions increment a counter and are blocked after the
threshold; empty actions trigger explicit feedback; forced next actions from later
monitors are executed only if still admissible and are discarded when the model
chooses a task-critical action itself. \\
\cmidrule(lr){2-3}
& \multirow{7}{*}{Trajectory Regulation Layer}
& \texttt{WorldModel} updates inventory, current location, visited/unvisited locations,
observed objects, target location, placed locations, destination locations, and
lamp location after every observation; subgoal advancement moves through FIND,
TAKE, CLEAN/HEAT/COOL, GOTO\_DEST, PUT, lamp, and EXAMINE states; post-step monitoring
catches empty turns, examine-without-lamp loops, oscillation, repeated open/close,
dead-end look/inventory loops, put/take ``nothing happens'', and generic stalls;
step guidance proposes a concrete admissible next action; budget checking warns when
search is late and forces PUT only when the held item and admissible put action
make completion deterministic. \\
\midrule
\multirow{24}{*}{\centering WebShop}
& \multirow{5}{*}{Environment Contract Layer}
& task parsing extracts item keywords, required color or color
alternative, size, material, max price, quantity, style, measurement specs, and
product category; page-type detection identifies home, search-result, and product
pages from observations and clickables; page-state tracking records selected
attributes, ASIN, current price, search queries, back-to-search loops, and page
stalls; tool-description patching tells the model to search with concise
keywords, click only visible options, select attributes before buying, and keep
budget words out of search text. \\
\cmidrule(lr){2-3}
& \multirow{5}{*}{Procedural Skill Layer}
& WebShop skill retrieval first filters skills by product category, then
BM25-ranks against the instruction; skills tell the agent that about 5\% over
budget can be acceptable, compound colors need exact-or-closest matching,
petite/tall/plus sizes are combined options, measurements can appear under size
or other attributes, food flavor and pack count are clickable variants, home
items use dimension options, code prefixes such as \texttt{01\# black} should be
matched by visible text, and unavailable exact values should fall back to the
closest option. \\
\cmidrule(lr){2-3}
& \multirow{5}{*}{Action Realization Layer}
& \texttt{pre\_validate\_action} accepts only WebShop search and click action syntax;
search queries are cleaned to remove price/budget suffixes;
\texttt{\_fuzzy\_match\_click} maps near-miss clicks to current clickables; unknown tools and
invisible click targets are blocked; repeated non-navigation clicks are blocked
after the threshold; \texttt{\_buy\_now\_precheck} blocks \texttt{click[buy now]} while
visible required color, size, spec, or unselected defensive attribute groups
remain; forced \texttt{click[buy now]} is cancelled if the precheck still fails. \\
\cmidrule(lr){2-3}
& \multirow{5}{*}{Trajectory Regulation Layer}
& post-step monitoring checks the first product-page title against the requested item
category; detects repeated back-to-search and duplicate-query loops; warns once
per ASIN when price exceeds budget beyond tolerance; detects product-page stalls
and asks for an immediate buy-or-back decision; step guidance proposes initial
searches, ranks visible search results, and builds an attribute checklist from
requirements and clickables; budget checking warns near the end and forces
\texttt{click[buy now]} only when the button is visible. \\
\bottomrule
\end{tabular}
\caption{AgentBench harness inventory under the four lifecycle layers (ALFWorld and WebShop).}
\label{tab:agentbench-inventory}
\end{table*}

\begin{table*}[t]
\huge
\centering
\scriptsize
\setlength{\tabcolsep}{2.2pt}
\renewcommand{\arraystretch}{0.86}
\begin{tabular}{>{\centering\arraybackslash}m{0.1\textwidth}>{\centering\arraybackslash}m{0.2\textwidth}p{0.7\textwidth}}
\toprule
Domain & Layer & Implemented harness content \\
\midrule
\multirow{27}{*}{\centering\mbox{OS Interaction}}
& \multirow{5}{*}{Environment Contract Layer}
& task parsing classifies the request as count-files, count-lines,
count-matches, count-unique, largest, smallest, list, read-content, system-info,
sum-size, average, mutate, or other; it extracts answer shape, target path,
extension, recursive/non-recursive scope, case-sensitivity, time filter, and size
filter; tool-description patching tells the model to call bash through the tool
interface, inspect paths before broad commands, use \texttt{find}/\texttt{grep}/
\texttt{awk}/\texttt{sort}/\texttt{wc} with correct counting semantics, avoid
dangerous commands, and submit only the requested answer. \\
\cmidrule(lr){2-3}
& \multirow{6}{*}{Procedural Skill Layer}
& OS skill retrieval tokenizes the task with stopword removal, filters by parsed
task type, applies a BM25 score threshold, and force-inserts high-confidence
skills such as case-insensitive matching; the skill library gives concrete command
patterns for file counts by extension/time/size, recursive search, excluding
directories, grep line counts, total line counts, unique fields/words, largest or
smallest files, size totals, disk/memory/process questions, output truncation,
bracketed/date grep, safe averages, hidden files, field extraction before
\texttt{sort -u}, non-recursive counts, IP/status parsing, max-number frequency,
atime versus mtime, LOC, and date-entry counts. \\
\cmidrule(lr){2-3}
& \multirow{5}{*}{Action Realization Layer}
& tool-call rescue converts malformed model text into bash or
\texttt{answer\_action} calls from JSON, keyword, positional, bare-command, and XML-like
formats; \texttt{pre\_validate\_action} rejects dangerous shell patterns; repeated identical
bash commands are blocked after the threshold; text-only loops trigger a forced
answer when a plausible candidate exists or finish when none exists; \texttt{normalize\_answer}
strips units/prose for integer or size answers while preserving the required
string/path answers; \texttt{note\_answer\_submitted} marks the task complete for later
monitors. \\
\cmidrule(lr){2-3}
& \multirow{7}{*}{Trajectory Regulation Layer}
& bash-state tracking records bash history, raw output, truncation, errors,
empty-output streaks, numeric candidates, string/path candidates, and implausible
values; post-step monitoring handles truncated output, command-not-found,
path/glob/permission errors, grep/find/xargs warnings, empty output, repeated
bash loops, text-only loops, \texttt{grep -c} per-file counts that must be
summed, post-\texttt{ls} false-zero cases, and budget force/warn; step guidance
lints missing recursion, wrong case handling, file-versus-line counting, bad
\texttt{find -path}, missing \texttt{tr} input, date extraction, and human-readable
size mistakes; it also aggregates extension component counts, formats
\texttt{wc} filename answers, verifies early numeric answers, and accepts correct
zeroes, promotes clean string/path candidates, and gives first-turn templates. \\
\midrule
\multirow{32}{*}{\centering DBBench}
& \multirow{6}{*}{Environment Contract Layer}
& task parsing classifies the SQL request as SELECT, INSERT, UPDATE,
DELETE, counting, ranking, MAX/MIN/SUM/AVG/COUNT aggregation, comparison, or
other; answer-shape detection decides scalar integer/float/string, single-column
multi-row, multi-column multi-row, or mutation-hash answer; schema-map building
mirrors DBBench identifier sanitization, 64-character truncation, and duplicate
suffixing; schema-card injection supplies the exact sanitized table and column names;
tool-description and system-prompt patching state MySQL
syntax, mutation, schema-card, and commit-format requirements. \\
\cmidrule(lr){2-3}
& \multirow{8}{*}{Procedural Skill Layer}
& DB skill retrieval filters skills by parsed SQL task type, BM25-ranks them,
and applies heuristic boosts for grouped answers, unique entities, and dataset-average
updates, next/previous row lookup, currency-text comparison, and insert requests;
skills give concrete reminders for backticking identifiers, MySQL
\texttt{CONCAT}, casting text numerics, DESCRIBE/SHOW recovery, LIKE/OR filters,
SELECT result shape, COUNT, ranking with ORDER BY/LIMIT, returning the requested
column not the rank, SUM/AVG denominators, INSERT value order, and all columns,
UPDATE/DELETE WHERE clauses, preview-then-mutate, multi-row commit formatting,
bare numeric commits, NULL-as-zero, mutation-before-commit, date passthrough,
GROUP BY/HAVING, BETWEEN, subqueries, apostrophe escaping, mutation verification,
and the exact inserted text formats. \\
\cmidrule(lr){2-3}
& \multirow{7}{*}{Action Realization Layer}
& tool-call rescue extracts \texttt{execute\_sql} or
\texttt{commit\_final\_answer} from XML, JSON, keyword syntax, and truncated
partial answers; automatic backtick repair wraps known sanitized table/column names while
leaving string literals untouched; dialect repair converts common SQLite-style
patterns to MySQL, where safe; dangerous SQL is blocked; commit gates prevent
mutation tasks from committing before INSERT/UPDATE/DELETE succeeds, prevent
empty non-mutation commits, and block scalar commits after clear multi-row
results; answer-list normalization formats scalar and multi-row answers; NULL
aggregate candidates become \texttt{0} when the evaluator expects zero; repeated
SQL with a plausible candidate can force \texttt{commit\_final\_answer}. \\
\cmidrule(lr){2-3}
& \multirow{7}{*}{Trajectory Regulation Layer}
& SQL-state tracking records raw and normalized SQL history, result text,
error kind, empty streak, candidate answer, answer shape, implausibility,
mutation-attempt flag, row/column count, and text-only streak; post-step monitoring
handles text-only tool loops, MySQL syntax errors, unknown columns, and unknown
tables, mutation tasks that only ran SELECT, NULL aggregates, repeated empty
mutation previews, repeated empty query results, identical SQL loops, and budget
force/warn; step guidance promotes plausible candidates, explains the tuple-string
format for multi-column rows, lints contains-without-LIKE, case-insensitive
without LOWER, and ranking without ORDER BY, supplies first-turn SQL templates,
and warns when numeric or scalar candidates look implausible. \\
\bottomrule
\end{tabular}
\caption{AgentBench harness inventory under the four lifecycle layers (OS Interaction and DBBench).}
\label{tab:agentbench-inventory-os-db}
\end{table*}

\section{Additional Experiments}
\label{app:exp}

\paragraph{Dataset statistics and splits.}
Table~\ref{tab:data_statistics} reports the data statistics used in our experiments. 
For each environment, we distinguish between the raw training pool, the subset used for harness evolution, and the held-out test/evaluation set. 
The ``Train Used'' column reports the size of the training subset available for harness evolution. Each evolution round uses a batch of up to 50 tasks drawn from this subset. 
This design keeps harness evolution efficient while ensuring that evaluation is performed on held-out tasks that are not used during harness construction. 
The ``Raw Pool Total'' column reports the full available pool size for each environment.

\paragraph{Detailed Evaluation Configuration}
Table~\ref{tab:eval-config} summarizes the detailed evaluation configuration. For the Procedural Skill Layer, we use only the top-1 retrieved skill across all experiments, preventing irrelevant skills from contaminating the model context.

\paragraph{Main Results.}
Table~\ref{tab:appendix-full-results} reports the full results of \method{} across 18 model backbones and 7 benchmarks, showing consistent gains across diverse models and environments.

\begin{table*}[t]
    \centering
    \scriptsize
    \setlength{\tabcolsep}{3.5pt}
    \resizebox{\textwidth}{!}{
    \begin{tabular}{lccccccccccccc}
    \toprule
    \multirow{2}{*}{Model}
    & \multicolumn{4}{c}{AgentBench}
    & \multicolumn{3}{c}{$\tau$-bench Airline}
    & \multicolumn{3}{c}{$\tau$-bench Retail}
    & \multicolumn{3}{c}{$\tau^2$-bench Telecom} \\
    \cmidrule(lr){2-5}
    \cmidrule(lr){6-8}
    \cmidrule(lr){9-11}
    \cmidrule(lr){12-14}
    & ALFWorld & WebShop & OS & DBBench
    & Pass@1 & Pass@3 & Pass$^3$
    & Pass@1 & Pass@3 & Pass$^3$
    & Pass@1 & Pass@3 & Pass$^3$ \\
    \midrule
    Qwen3-4B-Ins
    & 0.1651 & 0.2950 & 0.2150 & 0.4400
    & 0.3500 & 0.6500 & 0.0500
    & 0.4800 & 0.7000 & 0.2500
    & 0.2583 & 0.4750 & 0.0750 \\
    \quad w/ ~\method
    & \textbf{0.8807} & \textbf{0.4550} & \textbf{0.3960} & \textbf{0.6500}
    & \textbf{0.6000} & \textbf{0.7000} & \textbf{0.5000}
    & \textbf{0.6300} & \textbf{0.8000} & \textbf{0.4500}
    & \textbf{0.9000} & \textbf{1.0000} & \textbf{0.7750} \\
    \midrule
    Qwen3.5-4B
    & 0.4312 & 0.3450 & 0.4236 & 0.5933
    & 0.8500 & 0.9500 & 0.7000
    & 0.7917 & 0.9250 & 0.6500
    & 0.9750 & 1.0000 & 0.9250 \\
    \quad w/ ~\method
    & \textbf{0.9266} & \textbf{0.4150} & \textbf{0.4931} & \textbf{0.7133}
    & \textbf{0.8667} & \textbf{0.9500} & \textbf{0.7500}
    & \textbf{0.8167} & 0.8750 & \textbf{0.7000}
    & \textbf{1.0000} & \textbf{1.0000} & \textbf{1.0000} \\
    \midrule
    Qwen3.5-9B
    & 0.5688 & 0.3750 & 0.4028 & 0.6067
    & 0.8500 & 0.9000 & 0.7500
    & 0.8333 & 0.9250 & 0.6750
    & 0.9750 & 1.0000 & 0.9250 \\
    \quad w/ ~\method
    & \textbf{0.9174} & \textbf{0.4400} & \textbf{0.5069} & \textbf{0.7033}
    & \textbf{0.8833} & \textbf{0.9000} & \textbf{0.8500}
    & 0.8000 & 0.9000 & \textbf{0.7000}
    & 0.9667 & \textbf{1.0000} & \textbf{0.9250} \\
    \midrule
    Qwen2.5-7B-Ins
    & 0.1284 & 0.3000 & 0.2500 & 0.5133
    & 0.2000 & 0.4000 & 0.1000
    & 0.2333 & 0.4250 & 0.0750
    & 0.3500 & 0.6000 & 0.1500 \\
    \quad w/ ~\method
    & \textbf{0.8165} & \textbf{0.4450} & \textbf{0.3542} & \textbf{0.6833}
    & \textbf{0.4167} & \textbf{0.5500} & \textbf{0.3500}
    & \textbf{0.2833} & \textbf{0.5500} & \textbf{0.1000}
    & \textbf{0.5500} & \textbf{0.7750} & \textbf{0.2750} \\
    \midrule
    Qwen2.5-14B-Ins
    & 0.4036 & 0.3950 & 0.2917 & 0.5300
    & 0.1833 & 0.5000 & 0.0000
    & 0.4417 & 0.6750 & 0.2000
    & 0.4000 & 0.5750 & 0.2500 \\
    \quad w/ ~\method
    & \textbf{0.8349} & \textbf{0.4500} & \textbf{0.4097} & \textbf{0.6600}
    & \textbf{0.4833} & \textbf{0.6000} & \textbf{0.3500}
    & \textbf{0.6167} & \textbf{0.8750} & \textbf{0.3750}
    & \textbf{0.5833} & \textbf{0.8500} & \textbf{0.2750} \\
    \midrule
    Llama-3.1-8B-Ins
    & 0.0550 & 0.2250 & 0.3125 & 0.1733
    & 0.3000 & 0.4000 & 0.2500
    & 0.0750 & 0.1250 & 0.0250
    & 0.2417 & 0.3250 & 0.1500 \\
    \quad w/ ~\method
    & \textbf{0.8257} & \textbf{0.4250} & \textbf{0.3333} & \textbf{0.4967}
    & \textbf{0.3333} & \textbf{0.4000} & \textbf{0.3000}
    & 0.0667 & \textbf{0.1250} & 0.0000
    & \textbf{0.4250} & \textbf{0.5750} & \textbf{0.3500} \\
    \midrule
    xLAM-2-3B
    & 0.0275 & 0.1450 & 0.0694 & 0.2800
    & 0.2000 & 0.4000 & 0.0500
    & 0.4000 & 0.6750 & 0.2000
    & 0.2583 & 0.4500 & 0.1000 \\
    \quad w/ ~\method
    & \textbf{0.4312} & \textbf{0.3550} & \textbf{0.1111} & \textbf{0.5400}
    & \textbf{0.4000} & \textbf{0.4500} & \textbf{0.3000}
    & \textbf{0.5167} & 0.6500 & \textbf{0.3500}
    & \textbf{0.4667} & \textbf{0.8000} & \textbf{0.1500} \\
    \midrule
    Llama-xLAM-2-8B
    & 0.0092 & 0.1150 & 0.1389 & 0.2433
    & 0.3667 & 0.6500 & 0.1500
    & 0.6083 & 0.7750 & 0.4250
    & 0.3167 & 0.5750 & 0.1000 \\
    \quad w/ ~\method
    & \textbf{0.1193} & \textbf{0.4350} & \textbf{0.1597} & \textbf{0.6300}
    & \textbf{0.5500} & \textbf{0.6500} & \textbf{0.3500}
    & \textbf{0.6333} & \textbf{0.8750} & 0.3750
    & \textbf{0.4417} & \textbf{0.8000} & \textbf{0.1750} \\
    \midrule
    Qwen3-30B-A3B-Ins
    & 0.1101 & 0.3500 & 0.4444 & 0.5767
    & 0.4167 & 0.6000 & 0.2000
    & 0.4583 & 0.6750 & 0.2250
    & 0.3750 & 0.5750 & 0.1750 \\
    \quad w/ ~\method
    & \textbf{0.8807} & \textbf{0.4650} & \textbf{0.4583} & \textbf{0.7333}
    & \textbf{0.6000} & \textbf{0.7000} & \textbf{0.5000}
    & \textbf{0.5500} & \textbf{0.7250} & \textbf{0.3250}
    & \textbf{0.7083} & \textbf{0.8750} & \textbf{0.5250} \\
    \midrule
    Qwen3.5-27B
    & 0.8899 & 0.3850 & 0.5417 & 0.6833
    & 0.8833 & 0.9000 & 0.8500
    & 0.8333 & 0.9500 & 0.6750
    & 0.9417 & 1.0000 & 0.8750 \\
    \quad w/ ~\method
    & \textbf{0.8899} & \textbf{0.4600} & \textbf{0.5833} & \textbf{0.7400}
    & 0.8500 & \textbf{0.9000} & 0.8000
    & \textbf{0.8417} & 0.9250 & \textbf{0.7500}
    & \textbf{0.9833} & \textbf{1.0000} & \textbf{0.9500} \\
    \midrule
    Qwen3.5-35B-A3B
    & 0.7431 & 0.3800 & 0.5139 & 0.6133
    & 0.8167 & 1.0000 & 0.7000
    & 0.8417 & 1.0000 & 0.7000
    & 0.9500 & 1.0000 & 0.8750 \\
    \quad w/ ~\method
    & \textbf{0.9174} & \textbf{0.4300} & \textbf{0.5694} & \textbf{0.6933}
    & \textbf{0.8333} & 0.9500 & \textbf{0.7500}
    & 0.8250 & 0.9000 & \textbf{0.7250}
    & \textbf{0.9583} & \textbf{1.0000} & \textbf{0.9000} \\
    \midrule
    Qwen3.6-35B-A3B
    & 0.7890 & 0.3800 & 0.5208 & 0.6233
    & 0.8500 & 0.9000 & 0.7500
    & 0.8000 & 0.9000 & 0.6750
    & 0.9917 & 1.0000 & 0.9750 \\
    \quad w/ ~\method
    & \textbf{0.8991} & \textbf{0.4500} & \textbf{0.5417} & \textbf{0.7133}
    & \textbf{0.8833} & \textbf{1.0000} & \textbf{0.7500}
    & \textbf{0.8667} & \textbf{0.9750} & \textbf{0.7250}
    & \textbf{0.9917} & \textbf{1.0000} & \textbf{0.9750} \\
    \midrule
    Qwen3.6-27B
    & 0.9083 & 0.3850 & 0.5208 & 0.6833
    & 0.8333 & 0.9500 & 0.7000
    & 0.8417 & 0.9500 & 0.7250
    & 0.9667 & 1.0000 & 0.9000 \\
    \quad w/ ~\method
    & 0.8991 & \textbf{0.4300} & \textbf{0.5694} & \textbf{0.7433}
    & 0.8000 & \textbf{0.9500} & \textbf{0.7000}
    & 0.8333 & 0.9250 & 0.6750
    & \textbf{0.9750} & \textbf{1.0000} & \textbf{0.9500} \\
    \midrule
    Qwen2.5-32B-Ins
    & 0.7064 & 0.3650 & 0.3681 & 0.5533
    & 0.3333 & 0.4500 & 0.2000
    & 0.5333 & 0.9250 & 0.2000
    & 0.4500 & 0.5500 & 0.3000 \\
    \quad w/ ~\method
    & \textbf{0.9266} & \textbf{0.4650} & \textbf{0.4583} & \textbf{0.6933}
    & \textbf{0.5167} & \textbf{0.6500} & \textbf{0.4000}
    & \textbf{0.6417} & 0.8750 & \textbf{0.3500}
    & \textbf{0.7917} & \textbf{0.9250} & \textbf{0.6500} \\
    \midrule
    Qwen2.5-72B-Ins
    & 0.7890 & 0.3600 & 0.4444 & 0.5167
    & 0.4167 & 0.6500 & 0.2500
    & 0.6167 & 0.8750 & 0.3000
    & 0.4667 & 0.7000 & 0.1750 \\
    \quad w/ ~\method
    & \textbf{0.8532} & \textbf{0.4800} & \textbf{0.4722} & \textbf{0.7133}
    & \textbf{0.6034} & \textbf{0.7000} & \textbf{0.4500}
    & \textbf{0.7000} & 0.8500 & \textbf{0.4750}
    & \textbf{0.7167} & \textbf{1.0000} & \textbf{0.4000} \\
    \midrule
    Llama-3.3-70B-Ins
    & 0.1835 & 0.3800 & 0.3403 & 0.5833
    & 0.2667 & 0.3000 & 0.2500
    & 0.0667 & 0.1000 & 0.0500
    & 0.3083 & 0.3250 & 0.2750 \\
    \quad w/ ~\method
    & \textbf{0.9266} & \textbf{0.4400} & \textbf{0.3958} & \textbf{0.6867}
    & \textbf{0.3000} & \textbf{0.3500} & \textbf{0.2500}
    & \textbf{0.0750} & 0.0750 & \textbf{0.0750}
    & \textbf{0.3333} & \textbf{0.3750} & \textbf{0.3000} \\
    \midrule
    xLAM-2-32B
    & 0.3761 & 0.2300 & 0.2639 & 0.0533
    & 0.3833 & 0.6000 & 0.1500
    & 0.6250 & 0.8750 & 0.3750
    & 0.3917 & 0.6500 & 0.1250 \\
    \quad w/ ~\method
    & \textbf{0.9174} & \textbf{0.4600} & \textbf{0.3472} & \textbf{0.3567}
    & \textbf{0.7000} & \textbf{0.8500} & \textbf{0.5000}
    & \textbf{0.7083} & \textbf{0.8750} & \textbf{0.5000}
    & \textbf{0.4583} & \textbf{0.7750} & \textbf{0.1250} \\
    \midrule
    Llama-xLAM-2-70B
    & 0.1101 & 0.2400 & 0.1875 & 0.4433
    & 0.4500 & 0.6500 & 0.1500
    & 0.6333 & 0.8000 & 0.4000
    & 0.3333 & 0.5250 & 0.1250 \\
    \quad w/ ~\method
    & \textbf{0.6422} & \textbf{0.4200} & \textbf{0.2500} & \textbf{0.4833}
    & \textbf{0.6500} & \textbf{0.7500} & \textbf{0.5500}
    & \textbf{0.7250} & \textbf{0.9250} & \textbf{0.5000}
    & \textbf{0.4917} & \textbf{0.7750} & \textbf{0.2500} \\
    \bottomrule
    \end{tabular}
    }
    \caption{
    Full main results across all evaluated model backbones and benchmarks. For AgentBench environments, we report pass@1 scores. For $\tau$-bench and $\tau^2$-bench environments, we report Pass@1, Pass@3, and Pass$^3$. Each model is evaluated with and without \method. Bold values indicate that \method{} improves or matches the corresponding no-harness result.
    }
    \label{tab:appendix-full-results}
\end{table*}
\begin{table*}[!tp]
    \centering
    \fontsize{7}{8.5}\selectfont
    \setlength{\tabcolsep}{3pt}
    \begin{tabular*}{\textwidth}{@{\extracolsep{\fill}}llrrrr@{}}
    \toprule
    \textbf{Model} & \textbf{Benchmark} & \textbf{w/o} & \textbf{ALIGN} & \textbf{AgentSquare} & \textbf{Life-Harness} \\
    \midrule
    \multirow{2}{*}{Qwen3-4B-Ins} & ALFWorld & 16.51 & 28.36 & 29.36 & \textbf{88.07} \\
     & WebShop & 29.50 & 18.00 & 18.50 & \textbf{45.50} \\
    \midrule
    \multirow{2}{*}{Qwen3.5-4B} & ALFWorld & 43.12 & 26.32 & 33.03 & \textbf{92.66} \\
     & WebShop & 34.50 & 16.00 & 38.50 & \textbf{41.50} \\
    \midrule
    \multirow{2}{*}{Qwen3.5-9B} & ALFWorld & 56.88 & 12.69 & 30.28 & \textbf{91.74} \\
     & WebShop & 37.50 & 18.00 & 35.00 & \textbf{44.00} \\
    \midrule
    \multirow{2}{*}{Qwen2.5-7B-Ins} & ALFWorld & 12.84 & 58.96 & 32.11 & \textbf{81.65} \\
     & WebShop & 30.00 & 28.00 & 19.50 & \textbf{44.50} \\
    \midrule
    \multirow{2}{*}{Qwen2.5-14B-Ins} & ALFWorld & 40.36 & 58.21 & 30.28 & \textbf{83.49} \\
     & WebShop & 39.50 & 30.00 & 29.00 & \textbf{45.00} \\
    \midrule
    \multirow{2}{*}{Llama-3.1-8B-Ins} & ALFWorld & 5.50 & 12.69 & 0.00 & \textbf{82.57} \\
     & WebShop & 22.50 & 12.00 & 0.00 & \textbf{42.50} \\
    \midrule
    \multirow{2}{*}{xLAM-2-3B} & ALFWorld & 2.75 & 11.19 & 0.00 & \textbf{43.12} \\
     & WebShop & 14.50 & 0.00 & 4.00 & \textbf{35.50} \\
    \midrule
    \multirow{2}{*}{Llama-xLAM-2-8B} & ALFWorld & 0.92 & 3.73 & 0.00 & \textbf{11.93} \\
     & WebShop & 11.50 & 4.00 & 1.51 & \textbf{43.50} \\
    \midrule
    \multirow{2}{*}{Qwen3-30B-A3B-Ins} & ALFWorld & 11.01 & 38.06 & 43.12 & \textbf{88.07} \\
     & WebShop & 35.00 & 18.00 & 23.00 & \textbf{46.50} \\
    \midrule
    \multirow{2}{*}{Qwen3.5-27B} & ALFWorld & \textbf{88.99} & 64.93 & 67.89 & \textbf{88.99} \\
     & WebShop & 38.50 & 30.00 & 44.50 & \textbf{46.00} \\
    \midrule
    \multirow{2}{*}{Qwen3.5-35B-A3B} & ALFWorld & 74.31 & 62.69 & 64.22 & \textbf{91.74} \\
     & WebShop & 38.00 & 20.00 & \textbf{43.00} & \textbf{43.00} \\
    \midrule
    \multirow{2}{*}{Qwen3.6-35B-A3B} & ALFWorld & 78.90 & 73.88 & 60.55 & \textbf{89.91} \\
     & WebShop & 38.00 & 38.00 & 44.50 & \textbf{45.00} \\
    \midrule
    \multirow{2}{*}{Qwen3.6-27B} & ALFWorld & \textbf{90.83} & 77.61 & 73.39 & 89.91 \\
     & WebShop & 38.50 & 30.00 & \textbf{47.00} & 43.00 \\
    \midrule
    \multirow{2}{*}{Qwen2.5-32B-Ins} & ALFWorld & 70.64 & 71.64 & 27.52 & \textbf{92.66} \\
     & WebShop & 36.50 & 28.00 & 17.00 & \textbf{46.50} \\
    \midrule
    \multirow{2}{*}{Qwen2.5-72B-Ins} & ALFWorld & 78.90 & 74.63 & 50.46 & \textbf{85.32} \\
     & WebShop & 36.00 & 34.00 & 33.50 & \textbf{48.00} \\
    \midrule
    \multirow{2}{*}{Llama-3.3-70B-Ins} & ALFWorld & 18.35 & 56.72 & 47.71 & \textbf{92.66} \\
     & WebShop & 38.00 & 30.00 & 16.50 & \textbf{44.00} \\
    \midrule
    \multirow{2}{*}{xLAM-2-32B} & ALFWorld & 37.61 & 65.67 & 65.14 & \textbf{91.74} \\
     & WebShop & 23.00 & 18.00 & 25.50 & \textbf{46.00} \\
    \midrule
    \multirow{2}{*}{Llama-xLAM-2-70B} & ALFWorld & 11.01 & 18.66 & 26.61 & \textbf{64.22} \\
     & WebShop & 24.00 & 0.00 & \textbf{50.00} & 42.00 \\
    \bottomrule
    \end{tabular*}
    \caption{Comparison with ALIGN and AgentSquare across 18 model backbones on ALFWorld and WebShop. Accuracy is reported in percentages; bold denotes the highest accuracy in each row. ALIGN follows its released task split; AgentSquare uses the Life-Harness task selection.}
    \label{tab:align-agentsquare}
\end{table*}
\begin{table*}[!tp]
    \centering
    \fontsize{7}{8.5}\selectfont
    \setlength{\tabcolsep}{2pt}
    \begin{tabular*}{\textwidth}{@{\extracolsep{\fill}}lllrrrrrr@{}}
    \toprule
    \textbf{Benchmark} & \textbf{Method} & \textbf{Metric}
    & \textbf{Round 1} & \textbf{Round 2} & \textbf{Round 3}
    & \textbf{Round 4} & \textbf{Round 5} & \textbf{Round 6} \\
    \midrule
    \multirow{6}{*}{ALFWorld} & \multirow{2}{*}{Meta-Harness} & Accuracy & 32.00 & 34.00 & 36.00 & 36.00 & 36.00 & 38.00 \\
    & & Cost & 1,844,422 & 8,005,177 & 11,587,642 & 20,431,198 & 26,129,534 & 37,138,309 \\
     & \multirow{2}{*}{AHE} & Accuracy & 20.00 & 64.00 & 98.00 & 96.00 & 96.00 & 98.00 \\
    & & Cost & 19,811,543 & 34,404,126 & 48,259,083 & 61,652,816 & 83,687,531 & 95,632,347 \\
     & \multirow{2}{*}{Life-Harness} & Accuracy & 28.00 & 28.00 & 72.00 & 82.00 & 86.00 & 88.00 \\
    & & Cost & 4,161,138 & 12,190,452 & 14,076,471 & 19,234,043 & 23,082,260 & 24,288,432 \\
    \midrule
    \multirow{6}{*}{DBBench} & \multirow{2}{*}{Meta-Harness} & Accuracy & 50.00 & 52.00 & 54.00 & 54.00 & 50.00 & 50.00 \\
    & & Cost & 6,207,521 & 11,129,491 & 15,462,932 & 19,411,225 & 23,968,214 & 32,813,601 \\
     & \multirow{2}{*}{AHE} & Accuracy & 62.00 & 52.00 & 52.00 & 40.00 & 44.00 & 40.00 \\
    & & Cost & 22,526,526 & 39,553,218 & 59,377,453 & 74,165,373 & 89,109,728 & 104,524,554 \\
     & \multirow{2}{*}{Life-Harness} & Accuracy & 60.00 & 66.00 & 72.00 & 58.00 & 56.00 & 70.00 \\
    & & Cost & 1,440,875 & 3,114,184 & 5,352,370 & 7,673,629 & 10,090,936 & 11,204,889 \\
    \midrule
    \multirow{6}{*}{OS} & \multirow{2}{*}{Meta-Harness} & Accuracy & 54.00 & 54.00 & 40.00 & 38.00 & 44.00 & 44.00 \\
    & & Cost & 4,498,792 & 6,217,706 & 13,970,941 & 15,285,055 & 23,800,235 & 26,787,235 \\
     & \multirow{2}{*}{AHE} & Accuracy & 36.00 & 56.00 & 60.00 & 60.00 & 54.00 & 56.00 \\
    & & Cost & 9,228,392 & 20,312,020 & 34,483,700 & 52,001,022 & 68,599,918 & 93,994,363 \\
     & \multirow{2}{*}{Life-Harness} & Accuracy & 54.00 & 54.00 & 52.00 & 66.00 & 56.00 & 62.00 \\
    & & Cost & 2,155,531 & 4,468,800 & 8,748,913 & 14,528,703 & 19,816,520 & 25,709,768 \\
    \midrule
    \multirow{6}{*}{WebShop} & \multirow{2}{*}{Meta-Harness} & Accuracy & 40.00 & 48.00 & 48.00 & 46.00 & 42.00 & 38.00 \\
    & & Cost & 2,192,980 & 5,495,748 & 11,348,118 & 22,286,791 & 30,234,742 & 36,741,591 \\
     & \multirow{2}{*}{AHE} & Accuracy & 38.00 & 36.00 & 42.00 & 34.00 & 42.00 & 36.00 \\
    & & Cost & 19,530,428 & 31,618,208 & 52,015,559 & 65,935,271 & 79,875,431 & 103,843,620 \\
     & \multirow{2}{*}{Life-Harness} & Accuracy & 38.00 & 42.00 & 48.00 & 36.00 & 38.00 & 42.00 \\
    & & Cost & 6,778,022 & 10,291,263 & 18,581,490 & 26,987,334 & 32,180,838 & 37,750,993 \\
    \midrule
    \multirow{6}{*}{Airline} & \multirow{2}{*}{Meta-Harness} & Accuracy & 30.00 & 30.00 & 30.00 & 30.00 & 30.00 & 30.00 \\
    & & Cost & 2,622,885 & 7,835,642 & 15,899,971 & 25,881,103 & 39,360,135 & 53,088,352 \\
     & \multirow{2}{*}{AHE} & Accuracy & 10.00 & 30.00 & 45.00 & 41.67 & 41.67 & 41.67 \\
    & & Cost & 11,435,194 & 19,827,536 & 33,097,411 & 45,764,819 & 56,115,000 & 64,497,756 \\
     & \multirow{2}{*}{Life-Harness} & Accuracy & 56.67 & 56.67 & 55.00 & 58.33 & 55.00 & 60.00 \\
    & & Cost & 17,892,199 & 21,732,468 & 25,532,231 & 28,991,124 & 32,983,279 & 36,142,791 \\
    \midrule
    \multirow{6}{*}{Retail} & \multirow{2}{*}{Meta-Harness} & Accuracy & 32.00 & 38.00 & 38.00 & 48.00 & 36.00 & 40.00 \\
    & & Cost & 1,661,897 & 3,554,878 & 4,996,837 & 9,089,103 & 10,629,017 & 14,363,326 \\
     & \multirow{2}{*}{AHE} & Accuracy & 40.00 & 40.00 & 50.00 & 44.00 & 32.00 & 20.00 \\
    & & Cost & 17,352,611 & 28,084,458 & 41,489,740 & 69,778,855 & 81,253,916 & 93,141,101 \\
     & \multirow{2}{*}{Life-Harness} & Accuracy & 40.00 & 32.00 & 48.00 & 30.00 & 50.00 & 42.00 \\
    & & Cost & 3,725,478 & 7,254,878 & 10,072,270 & 20,196,616 & 22,707,923 & 25,706,750 \\
    \midrule
    \multirow{6}{*}{Telecom} & \multirow{2}{*}{Meta-Harness} & Accuracy & 30.00 & 32.00 & 28.00 & 28.00 & 28.00 & 30.00 \\
    & & Cost & 2,324,074 & 5,075,319 & 7,244,553 & 11,817,103 & 17,158,854 & 21,322,791 \\
     & \multirow{2}{*}{AHE} & Accuracy & 46.00 & 54.00 & 86.00 & 88.00 & 88.00 & 100.00 \\
    & & Cost & 18,510,187 & 35,043,410 & 49,187,608 & 63,274,049 & 75,883,376 & 88,348,598 \\
     & \multirow{2}{*}{Life-Harness} & Accuracy & 36.00 & 78.00 & 90.00 & 92.00 & 84.00 & 82.00 \\
    & & Cost & 11,711,806 & 16,036,978 & 23,454,801 & 30,082,366 & 37,984,775 & 42,638,396 \\
    \bottomrule
    \end{tabular*}
    \caption{Six-round evolution across seven benchmarks. Accuracy is measured on each round's training batch (\%); Cost is cumulative evolution LLM tokens, excluding solver and evaluation usage.}
    \label{tab:appendix-evolution-data}
\end{table*}

\begin{table}[!tp]
    \centering
    \fontsize{7}{8.5}\selectfont
    \setlength{\tabcolsep}{2pt}
    \begin{tabular*}{\linewidth}{@{\extracolsep{\fill}}lrr@{}}
    \toprule
    \textbf{Model} & \textbf{w/o} & \textbf{Life-Harness} \\
    \midrule
    Qwen3-4B-Ins & 1.75 & \textbf{5.26} \\
    Qwen3.5-4B & 36.84 & \textbf{43.86} \\
    Qwen3.5-9B & 61.40 & 61.40 \\
    Qwen2.5-7B-Ins & 3.51 & \textbf{5.26} \\
    Qwen2.5-14B-Ins & 15.79 & \textbf{21.05} \\
    Llama-3.1-8B-Ins & 3.51 & \textbf{5.26} \\
    xLAM-2-3B & 0.00 & 0.00 \\
    Llama-xLAM-2-8B & 0.00 & \textbf{3.51} \\
    Qwen3-30B-A3B-Ins & 14.04 & \textbf{17.54} \\
    Qwen3.5-27B & 73.68 & \textbf{80.70} \\
    Qwen3.5-35B-A3B & 66.67 & \textbf{73.68} \\
    Qwen3.6-35B-A3B & 54.39 & \textbf{73.68} \\
    Qwen3.6-27B & 70.18 & \textbf{82.46} \\
    Qwen2.5-32B-Ins & 15.10 & \textbf{20.40} \\
    Qwen2.5-72B-Ins & 8.77 & \textbf{17.54} \\
    Llama-3.3-70B-Ins & 14.04 & \textbf{15.79} \\
    xLAM-2-32B & 15.79 & \textbf{17.54} \\
    Llama-xLAM-2-70B & 3.51 & \textbf{8.77} \\
    \midrule
    Mean & 25.50 & \textbf{30.76} \\
    \bottomrule
    \end{tabular*}
    \caption{Full AppWorld Pass@1 results (\%) across 18 backbones. Bold denotes an accuracy improvement with Life-Harness.}
    \label{tab:appworld-full-results}
\end{table}

\section{AI Assistant Usage}
AI assistants were used for language editing and implementation support. The authors reviewed all AI-assisted content and code and took responsibility for the experimental design, analysis, and reported results.

\end{document}